 \documentclass[tablecaption=bottom,pmlr]{jmlr}

\usepackage{booktabs}
\usepackage[load-configurations=version-1]{siunitx} % newer version
\usepackage[utf8]{inputenc} % allow utf-8 input
\usepackage[T1]{fontenc}    % use 8-bit T1 fonts
\usepackage{hyperref}       % hyperlinks
\usepackage{url}            % simple URL typesetting
\usepackage{booktabs}       % professional-quality tables
\usepackage{amsfonts}       % blackboard math symbols
\usepackage{nicefrac}       % compact symbols for 1/2, etc.
\usepackage{microtype}      % microtypography

\usepackage{wrapfig}
\usepackage{amsmath}
\DeclareMathOperator*{\argmax}{arg\,max}

\usepackage{amssymb}
    
\usepackage{graphicx}

\usepackage[T1]{fontenc}
\usepackage{lmodern}

\theorembodyfont{\upshape}
\theoremheaderfont{\scshape}
\theorempostheader{:}
\theoremsep{\newline}

\jmlrvolume{148}
\jmlryear{2021}
\jmlrsubmitted{submission date}
\jmlrpublished{publication date}
\jmlrworkshop{NeurIPS 2020 Preregistration Workshop} % W&CP title

\title{On Initial Pools for Deep Active Learning}

\author{\Name{Akshay L Chandra}\thanks{Equal Contribution.} \Email{akshaychandra@iith.ac.in}\\
\Name{Sai Vikas Desai}$^{\textcolor{blue}{*}}$ \Email{cs17mtech11011@iith.ac.in}\\
\Name{Chaitanya Devaguptapu}$^{\textcolor{blue}{*}}$ \Email{cs19mtech110025@iith.ac.in}\\
\Name{Vineeth N Balasubramanian} \Email{vineethnb@cse.iith.ac.in}\\
\addr Department of Computer Science and Engineering \\
Indian Institute of Technology Hyderabad, India \\ \\}

\begin{document}

\maketitle

\begin{abstract}
Active Learning (AL) techniques aim to minimize the training data required to train a model for a given task. Pool-based AL techniques start with a small initial labeled pool and then iteratively pick batches of the most informative samples for labeling. Generally, the initial pool is sampled randomly and labeled to seed the AL iterations. While recent studies have focused on evaluating the robustness of various query functions in AL, little to no attention has been given to the design of the initial labeled pool for deep active learning. Given the recent successes of learning representations in self-supervised/unsupervised ways, we study if an \textit{intelligently sampled} initial labeled pool can improve deep AL performance. We investigate the effect of intelligently sampled initial labeled pools, including the use of self-supervised and unsupervised strategies, on deep AL methods. The setup, hypotheses, methodology, and implementation details were evaluated by peer review before experiments were conducted. Experimental results could not conclusively prove that intelligently sampled initial pools are better for AL than random initial pools in the long run, although a Variational Autoencoder-based initial pool sampling strategy showed interesting trends that merit deeper investigation.

% We describe our experimental details, implementation details, datasets, performance metrics as well as planned ablation studies in this proposal. If intelligently sampled initial pools improve AL performance, our work could make a positive contribution to boosting AL performance with no additional annotation, developing datasets with lesser annotation cost in general, and promoting further research in the use of unsupervised learning methods for AL. 

\end{abstract}


\section{Introduction}
\label{sec_intro}
With the success of convolutional neural networks (CNNs) in supervised learning on a wide range of tasks, several large and high-quality datasets have been developed. However, data annotation remains a key bottleneck for deep learning practitioners. Depending on the task, data annotation cost may vary from a few seconds to a few hours per sample and, in many real-world scenarios, supervision of domain experts is necessary (\cite{Bearman2016WhatsTP}). Active learning (AL) methods aim to alleviate the data annotation bottleneck by labeling only a subset of the most informative samples from a large pool of unlabeled data. Various query methods (\cite{DBAL_gal2017deep, coreset_sener2018active, Ensembles_Beluch2018ThePO, yoo2019learning_loss_for_AL,VAAL_sinha2019variational, Mottaghi2019AdversarialRA}) have been recently proposed for AL in the context of deep neural network (DNN) models (deep AL). The problem is important in deep AL, since DNN models require large amounts of labeled data to learn. In this work, we focus on the popular pool-based AL framework, which starts with a small initial labeled pool after which AL is performed in multiple \textit{sample-label-train} cycles.

AL has been well-explored in the context of traditional (shallow) machine learning methods (\cite{settles2009active}).  Generally, before starting the AL cycles, a small randomly chosen subset of a dataset (with size around 1-10\% of the entire dataset), typically called the \textit{initial pool}, is labeled first to train an initial model.
Across all AL efforts so far, to the best of our knowledge, the initial pool is always sampled randomly and labeled (\cite{DBAL_gal2017deep, coreset_sener2018active,  Ensembles_Beluch2018ThePO, yoo2019learning_loss_for_AL, VAAL_sinha2019variational, uncertainty_lewis1994sequential}). 
This initial pool design strategy has generally worked well for AL in traditional/shallow ML models. However, the success of AL in DNNs has not been convincing yet, especially when such models are trained on large-scale datasets. On one hand, while there have been several encouraging newly proposed deep AL methods, deeper analysis of those methods in (\cite{Munjal2020TowardsRA, Mittal2019PartingWI, Lowell2019PracticalOT}) show that AL struggles to outperform random sampling baselines when slight changes are made to either datasets (class-imbalance) or training procedures (data augmentation, regularization, etc.). Interestingly, to the best of our knowledge, the design of better initial labeled pools received no attention by the deep AL community. Considering the tremendous success of self-supervised learning methods in recent years (\cite{simlcr_Chen2020ASF, rotation_Gidaris2018, deepcluster_Caron2018, inpainting_Pathak2016, count_Noroozi2017, jigsaw_Noroozi2016, context_Doersch2015}), we ask the question if choosing an initial labeled pool intelligently can improve AL performance.

In our work, we propose to perform an empirical study of deep AL methods while using initial labeled pools, sampled using methods other than random sampling. To investigate the effect of \textit{intelligently} sampled initial labeled pools on deep AL methods, we propose two sampling techniques, leveraging  state-of-the-art self-supervised learning methods and well-known clustering methods. In particular, we propose the following ways of choosing the initial pool:
\begin{itemize}
    \item Sample datapoints that a state-of-the-art self-supervised model finds \textit{challenging}, as observed using the trained model's loss on the data. 
    \item Cluster the unlabeled pool first and then perform sampling across each cluster. Equal proportions of datapoints are sampled from each cluster to make sure the chosen samples span the entire dataset.

\end{itemize}

We hypothesize that AL methods (we focus our efforts on deep AL methods) can benefit from more intelligently chosen initial pools, thus eventually reducing annotation cost in creation of datasets. Our empirical study will seek to address the following specific questions:
\begin{itemize}
    \item Can pool-based deep AL methods leverage design of intelligently sampled initial pools to improve AL performance?
    \item Can we exploit latest advancements in self-supervised learning to boost deep AL performance with no additional labeling cost?
    \item  Are some initial pools better than others? What makes an initial pool \textit{good}?
\end{itemize}

In a realistic training setting with measures to avoid overfitting (\textit{i.e.} regularization, batch norm, early stopping), we hypothesize that the generalization error of AL models starting with our initial pools will be lower than those of AL models starting with random initial pools, across AL cycles. However, as AL cycles increase, we expect to see shorter margins of error difference as the effect of our initial pools on the model performance could diminish with increase in labeled pool size. Studying the use of unsupervised/self-supervised learning in later epochs could be an interesting direction of future work. If initial pools do contribute to better model performances, our work could make a positive contribution to: (i) boosting AL performance with no additional annotation; (ii) developing datasets with lesser annotation cost in general; and (iii) promoting further research in the use of unsupervised learning methods for AL. On the other hand, if random initial pools perform better than our initial pools, the community will still have useful insights about this rather unexplored part of AL through this study.

% \vspace{-3mm}
\section{Related Work}

\noindent \textbf{Analysis of Active Learning: }In recent years, previous works have evaluated the robustness and effectiveness of deep AL methods for various tasks. \cite{Lowell2019PracticalOT} first reported some obstacles of deploying AL in practice by empirically evaluating consistency of AL gains over random sampling and transferability of active samples across models. Along the same lines, \cite{Mittal2019PartingWI} evaluated the performance of deep AL methods under data augmentation, low-budget regime and a label-intensive task of semantic segmentation. More recently,  \cite{Munjal2020TowardsRA} comprehensively tested the performance variance of deep AL methods across 25 runs of experiments. They considered various settings such as regularization, noisy oracles, varying annotation and validation set size, heavy data augmentation and class imbalance. However, none of these efforts have considered varying the sampling strategy for the initial pool.

\vspace{3mm}
\noindent \textbf{Exploiting Unlabeled Data: }Our focus is on finding out if initial pools with certain \textit{desirable} qualities can bolster AL performance. We exploit self-supervised pretext tasks to sample the initial pool more intelligently. Previous works have successfully managed to integrate unlabeled data into AL using self-supervised learning and semi-supervised learning. \cite{Simoni2019RethinkingDA} showed that initializing the target model with the features obtained from self-supervised pretraining gives AL a kick-start in performance. Contemporaneously, \cite{Mottaghi2019AdversarialRA} also used this technique in combination with a GAN based AL method and reported SOTA results on SVHN, CIFAR-10, ImageNet, CelebA datasets. This is enough evidence that exploiting self-supervised learning methods can boost AL performance, but the cited works operate in the model weight space. The importance of good initialization in weight space (\cite{good_init_Mishkin2016, 10.1109/ICCV.2015.123, pmlr-v9-glorot10a}) is well understood by the deep learning community; To the best of our knowledge, there have been no efforts in understanding the importance of good initialization in data space for deep AL methods. In case of traditional AL (before deep learning's popularity), there have been handful of encouraging works that support our hypothesis (\cite{Ka2004UCBS, Hu2010OffTA}). Both these works use \textit{k}-means clustering to initialize the initial label pool, \textit{k}-nearest neighbor algorithm for training and report better AL performance on small scale text classification tasks.  

\vspace{3mm}
\noindent \textbf{Model Loss for AL: }In our work, we use a trained model's loss to identify the most \textit{informative} unlabeled samples. Existing AL methods largely rely on using the target model's loss for active sampling. \cite{settles-expected-gradient-length} first proposed an AL framework by calculating Expected Gradient Length (EGL) where the learner queries an unlabeled instance which, if labeled and added to the labeled pool, would result in the new training gradient of the largest magnitude. More recently, \cite{yoo2019learning_loss_for_AL} proposed a loss prediction module which is attached to the target network to predict the loss value of unlabeled samples. In contrast to these methods, we strictly rely on a self-supervised model's loss, instead of the target model, since the initial pool needs to be selected/sampled, before any model is trained on the target data.      

% \vspace{-3mm}
\section{Methods and Experimental Protocol}
In this section, we describe: (i) the notations and setting for pool-based AL cycles; (ii) our strategies for sampling the initial pool; and (iii) the AL methods that are subsequently used to build on top of the initial labeled pool. Implementation details and other considered additional experiments and ablation studies are mentioned at the end of this section.

\subsection{Pool-based Active Learning Setting}
Given a dataset $\mathcal{D}$, we split it into train ($T_r$), validation ($V$), and test ($T_s$) sets. At the beginning, the train set is also treated as an unlabeled ($U$) set, from which samples are moved to a labeled set ($L$) after every AL cycle. Pool-based AL cycles operate on a set of labeled data $L_{0}$=$\{(x_i, y_i)\}_{i=1}^{N_L}$ and a large pool of unlabeled data $U_{0}$=$\{ x_i\}_{i=1}^{N_U}$, and model $\Phi_0$ is trained on $L_{0}$ in every AL cycle. In our setting, given $L_{0} = \emptyset$ to start with, a sampling function $\Psi(L_{0}, U_{0}, \Phi_0)$ parses $x_i \in U_0$, and selects $k$ (budget size) \textbf{samples}. These samples are then labeled by an \textbf{oracle} and added to $L_{0}$, resulting in a new, extended $L_{1}$ labeled set, which is then used to \textbf{retrain} $\Phi$. This cycle of \textbf{sample-label-train} repeats until the sampling budget is exhausted or a satisfying performance metric is achieved. In our case, we populate $L_{0}$ using our proposed methods, discussed in Section \ref{sec:our-methods}. Sec \ref{sec:al-methods} describes the query methods we use to perform the traditional AL cycles after this initial pool selection. We can confirm that there exists a good initial pool if the generalization error of models starting with our initial pools is lower than those of models starting with random initial pools across the AL cycles. 
\begin{figure}[h]
\centering
\begin{tabular}{ccc}
\includegraphics[width=3.2cm]{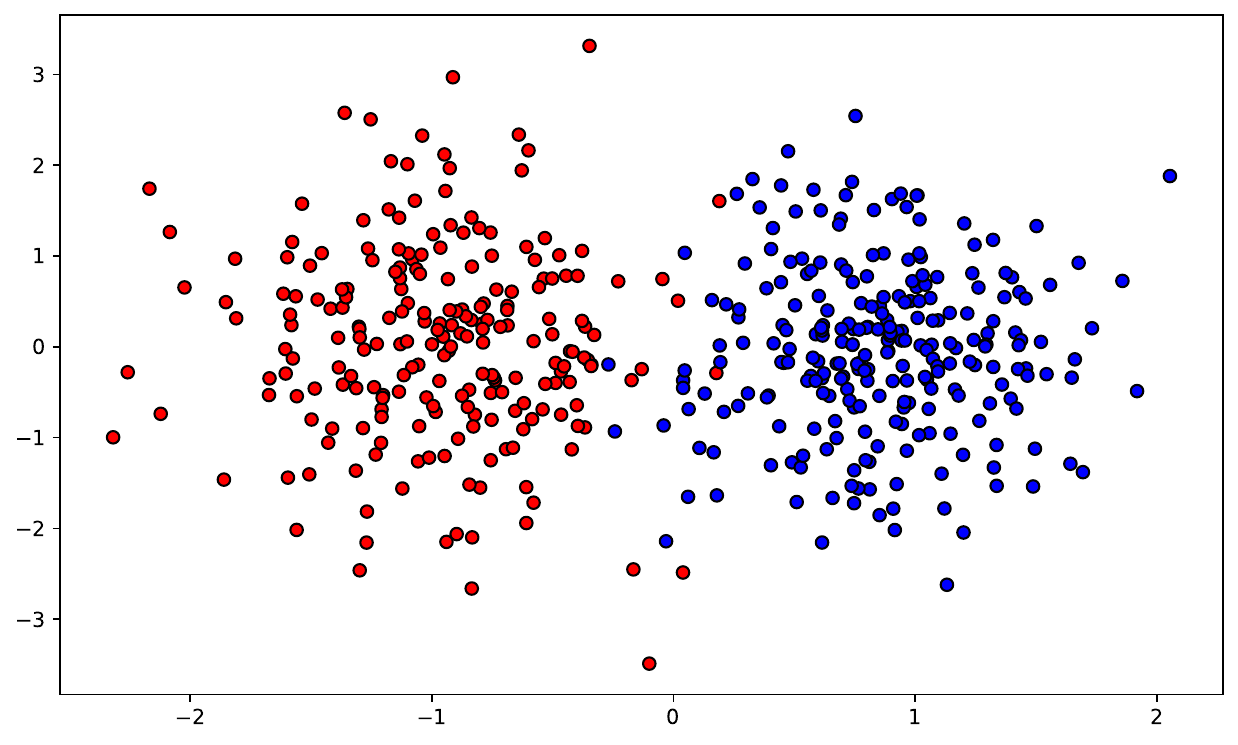}&
\includegraphics[width=3.5cm]{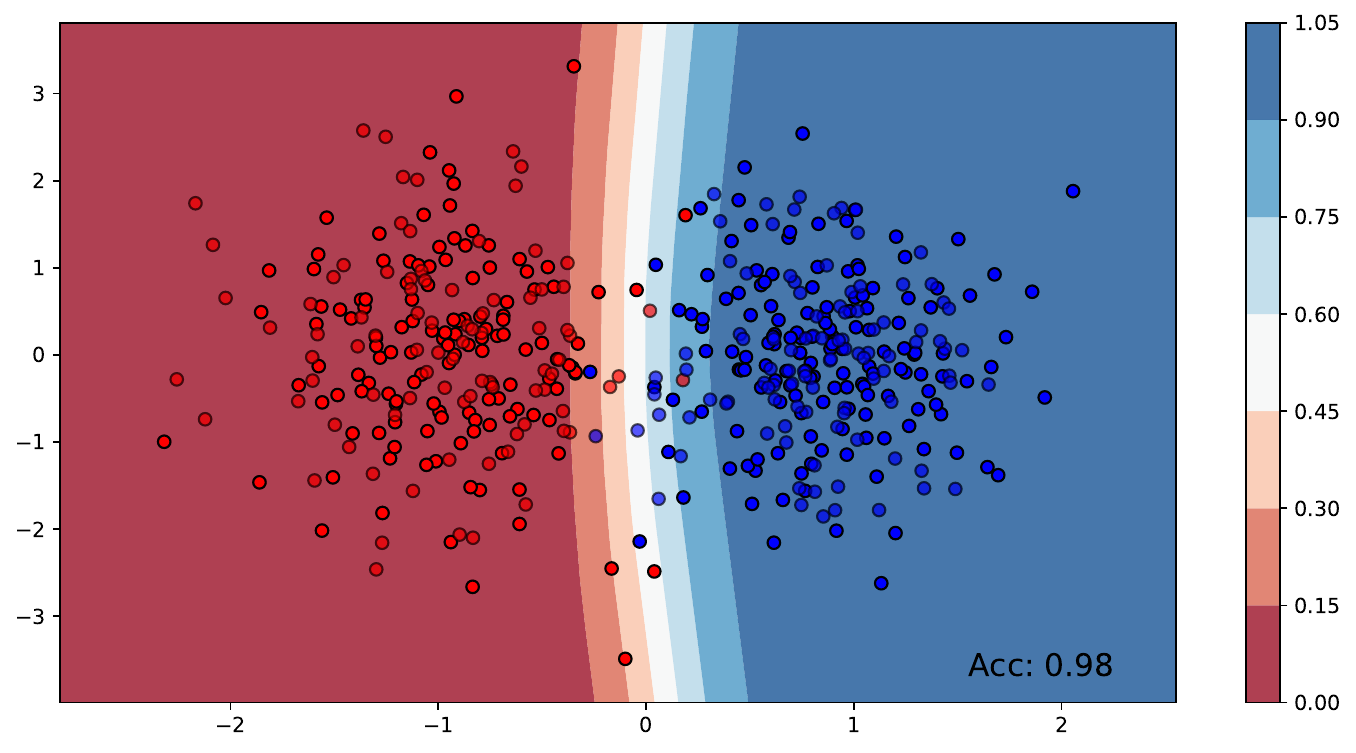}&
\includegraphics[width=3.5cm]{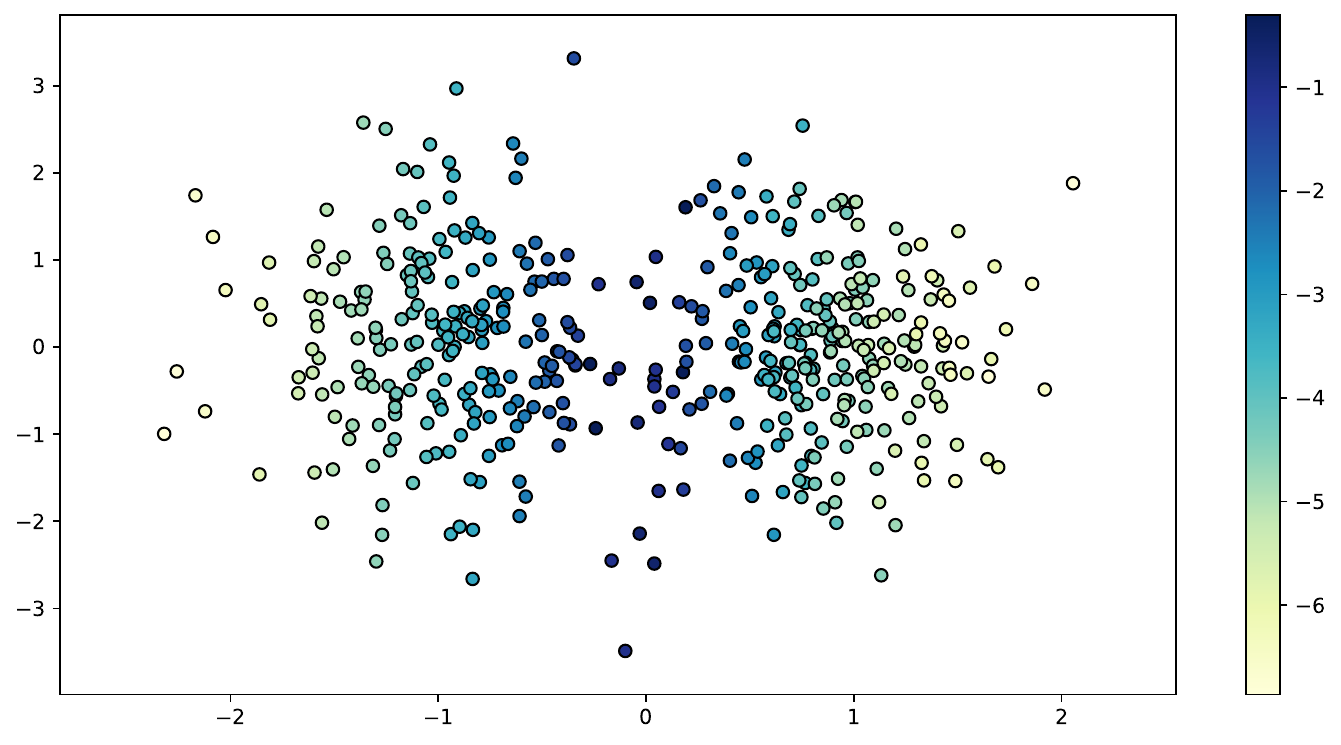}\\

(a)&(b)&(c) \\
\end{tabular}
\begin{tabular}{ccc}
\includegraphics[width=3.5cm]{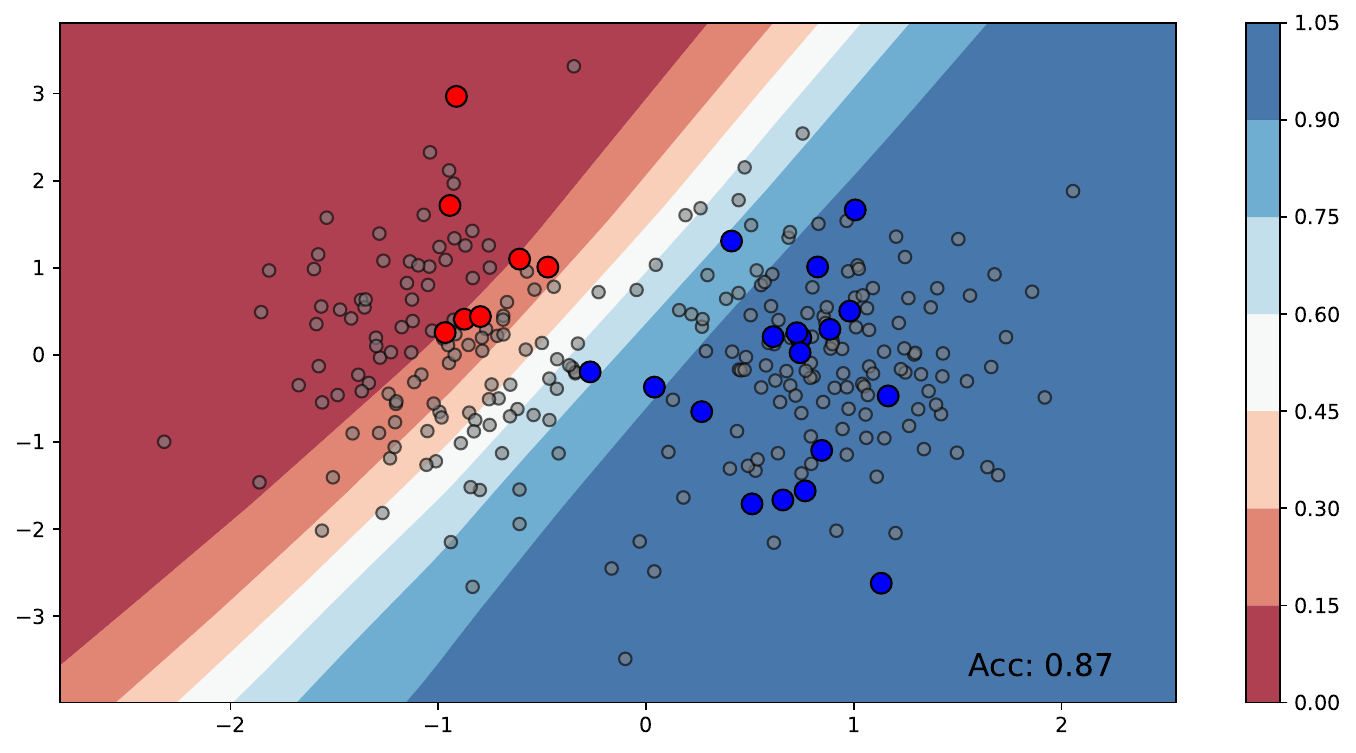}&
\includegraphics[width=3.5cm]{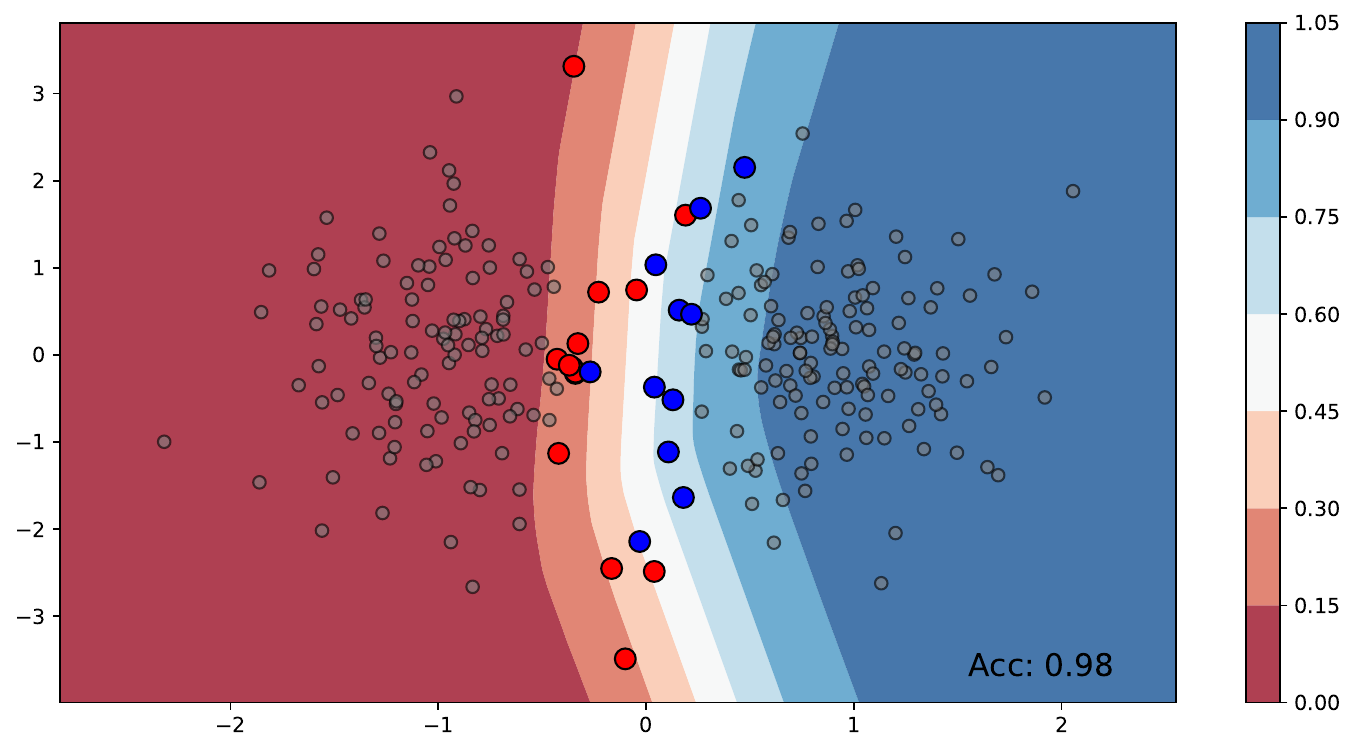}&
\includegraphics[width=3.5cm]{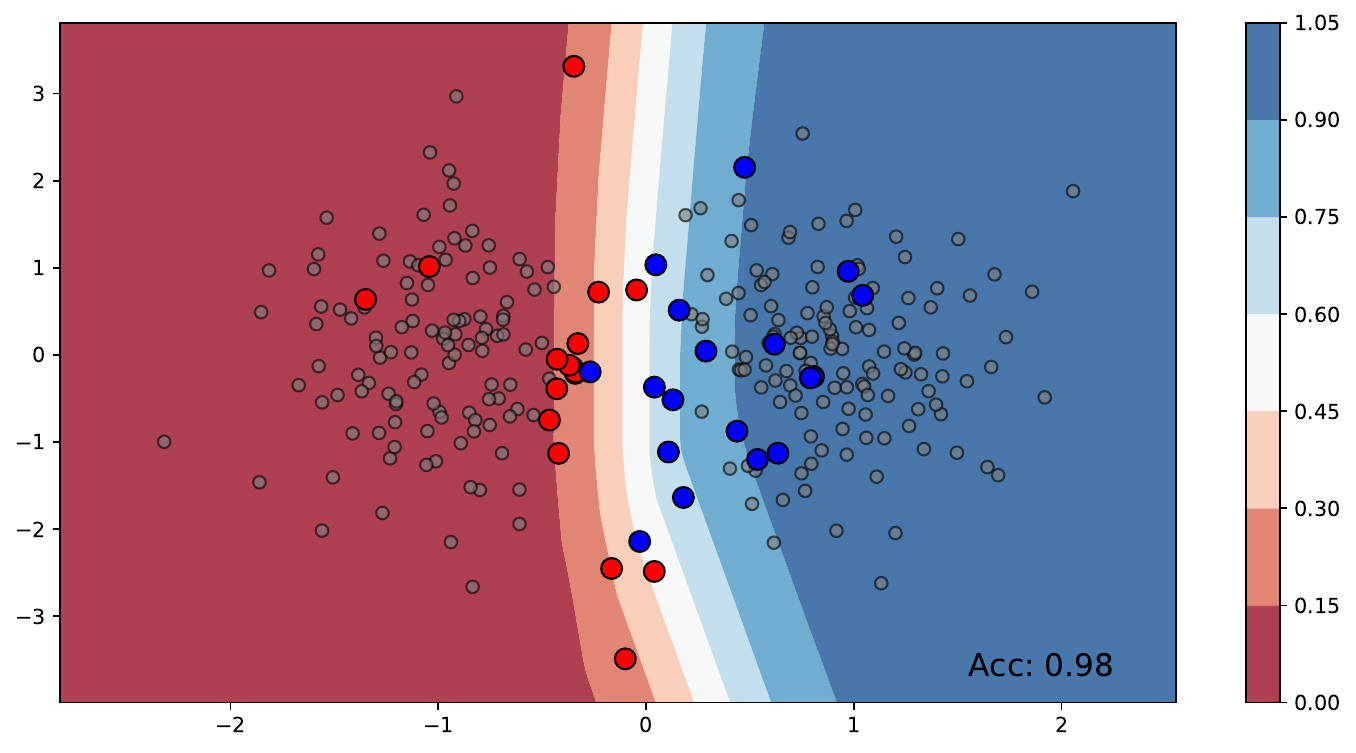}\\

(d)&(e)&(f) \\
\end{tabular}

\caption{Illustration of query strategies in traditional pool-based AL: (a) A toy dataset of 500 instances, evenly sampled from two class Gaussians; (b) Decision boundary of a logistic regression model trained on the dataset; (c) Trained model's \textit{CrossEntropy} loss on training instances; Decision boundary of logistic regression models trained on 35 instances chosen (d)  \textit{randomly} (e) using Least Confidence method \cite{uncertainty_lewis1994sequential} (f) using Max-Entropy method \cite{shannon-entropy}; Best viewed in color. Labeled instances are emphasized for clarity.}
\label{fig_al_example}
\end{figure}
 
\subsection{Proposed Initial Pool Sampling Strategies}\label{sec:our-methods}
We now describe our initial pool sampling strategies, which were briefly stated in Section \ref{sec_intro}. 

Our methods are fundamentally motivated by the hypothesis that samples considered \textit{challenging} for an unsupervised/self-supervised setting can help bootstrap AL methods through a more intelligent, well-guided choice of an initial labeled pool.   

\subsubsection{Self-Supervision Methods}\label{sec:selfsupervised}
As shown in Figure \ref{fig_al_example}, well-known pool-based AL methods rely on choosing samples with a high uncertainty of the target model trained on an initial labeled pool.

Since our focus is on choosing an initial labeled pool where there is no target model, we cannot use these AL query methods since we do not have access to labels to calculate the supervised model's loss on training data.
We hence train a self-supervised model on the entire unlabeled pool and identify samples as "challenging", where the self-supervised model's loss is relatively higher than that of others. The recent success of self-supervised learning in learning useful data representations (\cite{context_Doersch2015,jigsaw_Noroozi2016, count_Noroozi2017, rotation_Gidaris2018, deepcluster_Caron2018, simlcr_Chen2020ASF}) motivates us to hypothesize that such a model could help us sample the most \textit{informative} datapoints without any supervision.
    
Let $\tau$ be any self-supervised task with an objective to minimize the loss function $\mathcal{L}$. Let $\theta$ be trained weights obtained by solving $\tau$ on the unlabeled data pool $U$. We want the oracle to label and populate the initial pool $L_0$ with datapoints sampled by solving:
\begin{equation}
    \argmax_{i} \mathcal{L}^{\tau}(x_i; \theta) \quad \forall x_i \in U
\label{eq:loss-based}
\end{equation}
Since we are working with a learned model's loss, any self-supervised task can be used, making our proposed method task-agnostic. We have chosen tasks that are simple and easy to interpret, such as image inpainting (\cite{inpainting_Pathak2016}) and image rotation prediction (\cite{rotation_Gidaris2018}). For example, in case of the rotation prediction task, our strategy can be summarized as: \textit{if a trained rotation predictor struggles to rightly predict the rotation of a sample, even after looking at it while training, then it is a hard sample - thus human labeling is needed}. In addition to the above tasks, we will also train a Variational Autoencoder (VAE) (\cite{Kingma2014AutoEncodingVB}) as one of our tasks where datapoints with highest loss \textit{i.e.} hard to reconstruct images are sampled for the initial pool. We want to do this to understand how complexity of self supervised tasks (\textit{e.g.} image inpainting task is more complex than VAEs) relates to efficiency of the sampled initial pool using those tasks.

\subsubsection{Unsupervised Learning (Clustering) Methods}\label{sec:unsupervised}
Sampling bias is the most fundamental challenge posed by AL especially in case of uncertainty based AL methods (\cite{sampling_bias_Dasgupta2011TwoFO}).
Assume AL is performed on a dataset with data distribution $\mathcal{D}$. But as AL cycles proceed, and datapoints are sampled and labeled based on increasingly confident assessments of their informativeness, the labeled set starts to look less like $\mathcal{D}$. This problem is further exacerbated by highly imbalanced real-world datasets where random initial samples, with high probability, may not span the entire data distribution $\mathcal{D}.$
To overcome this, several works proposed diversity based methods (\cite{coreset_sener2018active, VAAL_sinha2019variational}) whose fundamental goal is to sample unlabeled datapoints from a non-sampled area of $\mathcal{D}$ such that all areas of $\mathcal{D}$ are seen by the target model. Motivated by these methods and their success, we propose a clustering-based sampling method for choosing the initial pool such that the sampled points spans all area of $\mathcal{D}$ (\textit{i.e.}, all clusters) even before AL starts. In a way, this is analogous to exploration in AL (\cite{Bondu2010ExplorationVE}), albeit in an unsupervised way.

We assume that number of classes to be labeled ($K$) in the dataset $\mathcal{D}$ is known apriori. If a clustering algorithm is applied on the unlabeled data $U$=$\{ x_i\}_{i=1}^{N_U}$ to obtain $K$ clusters and every datapoint $x_i$ is assigned only one cluster $C_j$, we get $K$ disjoint sample sets $C$ = $\{C_1, C_2, ..., C_K\}$. If the initial pool budget is $B$ samples, we sample $\frac{B}{K}$ datapoints from each cluster. Equal weight is given to each cluster to make sure initial pool is populated with datapoints that span the entire $\mathcal{D}$. As another variant to this method, we will also experiment by sampling $\frac{|C_j| * B}{N_U}$ datapoints from each cluster, keeping the original cluster proportions intact. We will use DeepCluster (\cite{deepcluster_Caron2018}) and $k$-means as clustering methods in our experiments.

\vspace{-3mm}
\subsection{Active Learning Query Methods}\label{sec:al-methods}
In order to study the usefulness of the choice of the initial labeled pool across AL methods, we need to study different AL query methods in later cycles of model updation. Modern pool-based AL methods may be broadly classified into three categories. We will evaluate the effectiveness of our sampling methods on AL methods from all three categories: 
\begin{itemize}
    \item \textbf{Uncertainty Sampling:} Least Confidence (LC) (\cite{uncertainty_lewis1994sequential}), Max-Entropy (ME) (\cite{shannon-entropy}), Min-Margin (MM) (\cite{margin_Scheffer01activehidden}) \& Deep Bayesian AL (DBAL) (\cite{DBAL_gal2017deep})
    \item \textbf{Diversity Sampling:} Coreset (greedy) (\cite{coreset_sener2018active}) \& Variational Adversarial AL (VAAL) (\cite{VAAL_sinha2019variational})
    \item \textbf{Query-by-Committee Sampling:} Ensemble with Variation Ratio (ENS-varR) (\cite{Ensembles_Beluch2018ThePO}) (3 ResNet18 models) \& ensemble variants of Least Confidence (ENS-LC), Max-Entropy (ENS-ME) and Margin Sampling (ENS-MM) 
\end{itemize}

All the above methods are already implemented in the AL toolkit offered by \cite{Munjal2020TowardsRA}, and we will leverage it to study the methods. 

\section{Implementation Details}
Following recent deep AL efforts, we will use MNIST, CIFAR-10, CIFAR-100 and Tiny ImageNet-200 (\cite{tinyimagenet_Le2015}) datasets in our experiments. We use the AL methods, model architectures, data augmentation schemes and implementation details from \cite{Munjal2020TowardsRA} for our experiments.

For all datasets, we plan to tune hyperparameters using grid search. However, going by previous works, we expect to use an Adam optimizer (\cite{adam_DBLP}) across the datasets. For datasets CIFAR-10 and CIFAR-100, we expect to use learning rate (\textit{lr}), weight decay (\textit{wd}) from \cite{Munjal2020TowardsRA} - ($lr=5e^{-4}$, $wd=5e^{-4}$) and ($lr=5e^{-4}$ and $wd=0$) respectively. For all datasets, we augment the data with random horizontal flips ($p$ = $0.5$) and normalize them using statistics provided in \footnote{\url{https://github.com/pytorch/examples/}}, \footnote{\url{https://github.com/kuangliu/pytorch-cifar}}. We will use ResNet18 (\cite{resnet_he2016deep}) for all our experiments.

\vspace{3mm}
\noindent \textbf{AL Details: }As usually done in most AL work, we will initialize $L_0$ with 10\% of the unlabeled set $U$ and in every AL cycle 10\% of the original unlabeled set $U$ will be sampled, labeled and moved to labeled set $L_i$. However, we expect some changes in AL cycle details due to irregularities between datasets (\textit{e.g.} MNIST is easier to learn compared to Tiny ImageNet) and those changes will be reported appropriately post-experiments.

\vspace{3mm}
\noindent \textbf{Performance Metrics:} We will measure accuracy on the test set after every AL cycle (including after the choice of the initial labeled pool). Our initial pool sampling strategies will be compared against a random selection of the initial pool (the default option used today), and all our results will be reported (as mean and std) over 5 trials to avoid any randomness bias in the results. 

We also plan to visualize the chosen initial labeled pool using t-SNE embeddings in case this provides any understanding of sampling strategies that work best. We will also examine overlap between every labeled pool acquired during all AL cycles when our initial pool sampling strategy is used against a random choice.  This would allow us to know if initial pool played any role in altering the labeled pools (for better or worse). 

\subsection{Additional Experiments}\label{sec:additional-exp}
In practice, populating the initial pool only with \textit{challenging} datapoints may not be fully conducive for learning. Hence, we plan to follow  \cite{Roy2018DeepAL} and split the sorted list obtained by solving Eqn (\ref{eq:loss-based}) into $n$ equal-sized bins. If the initial pool budget size is $B$, we query $\frac{B}{n}$ highest scored images from the top $(n-1)$ bins (hard samples) and $\frac{B}{n}$ lowest scored images from the last bin (easy samples). So the resultant batch contains images from different regions of the score space. In the experiments, we will use 2, 5 and 10 as the values of $n$.

Additionally, we will test the usefulness of our sampling methods on AL for imbalanced data. For this, we will follow \cite{class_imbalance_Cui2019} to simulate a long-tailed distribution of classes on CIFAR-10, by following power law. 

\begin{figure}
\begin{tabular}{cccc}
\includegraphics[width=0.22\linewidth]{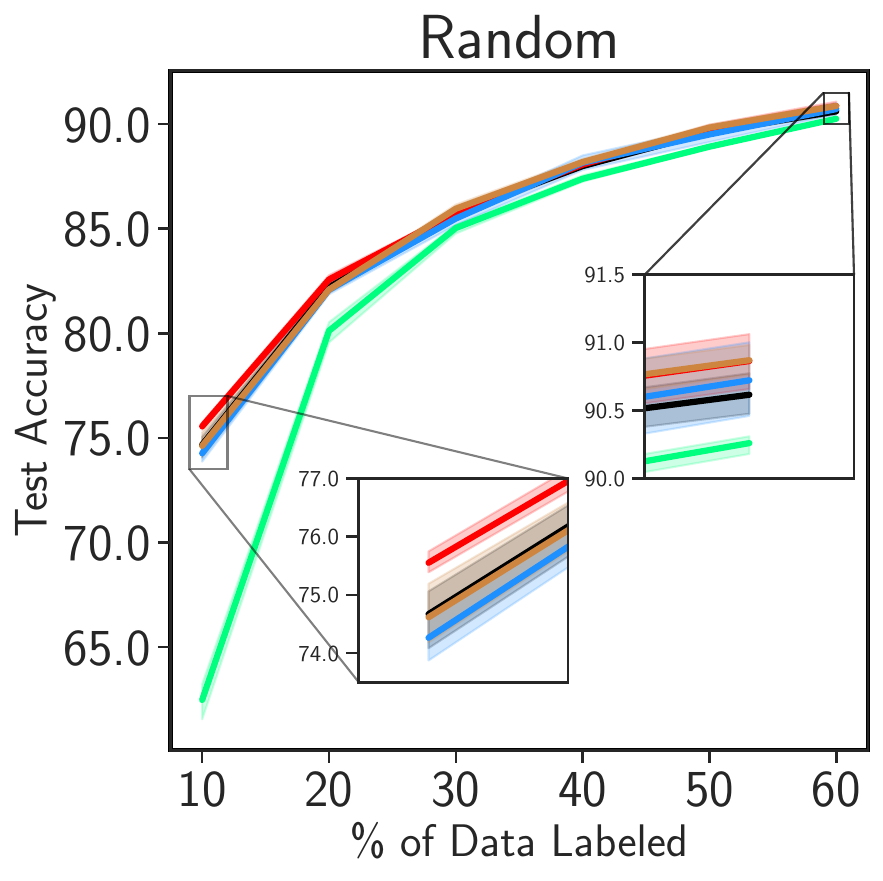}&
\includegraphics[width=0.22\linewidth]{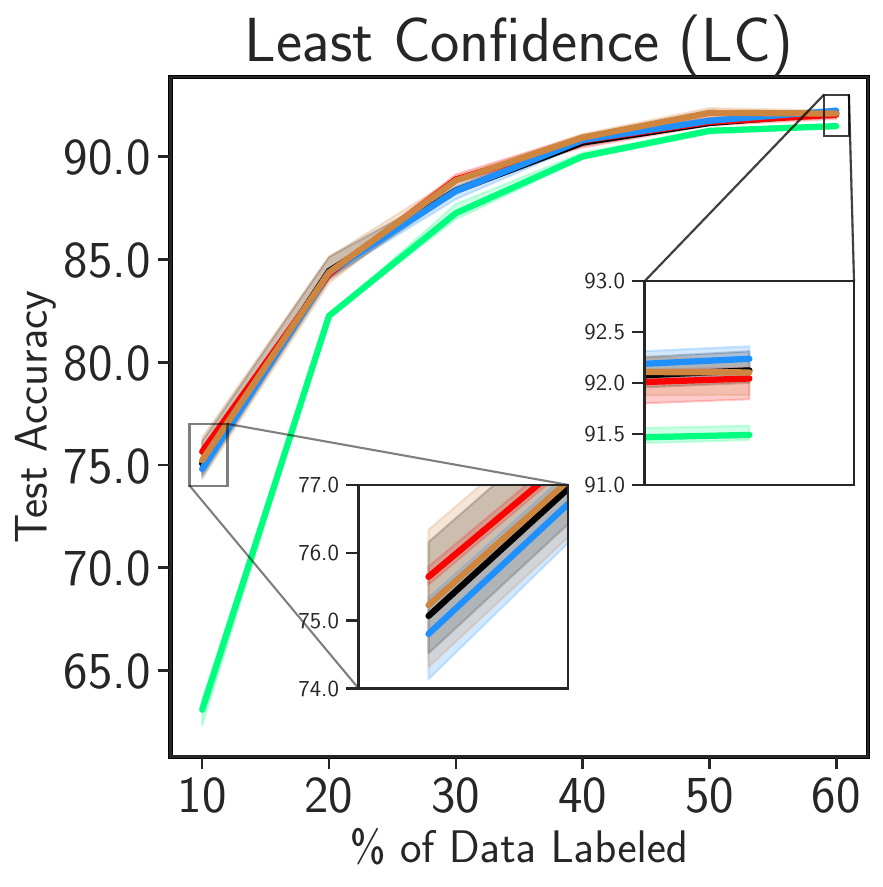}&
\includegraphics[width=0.22\linewidth]{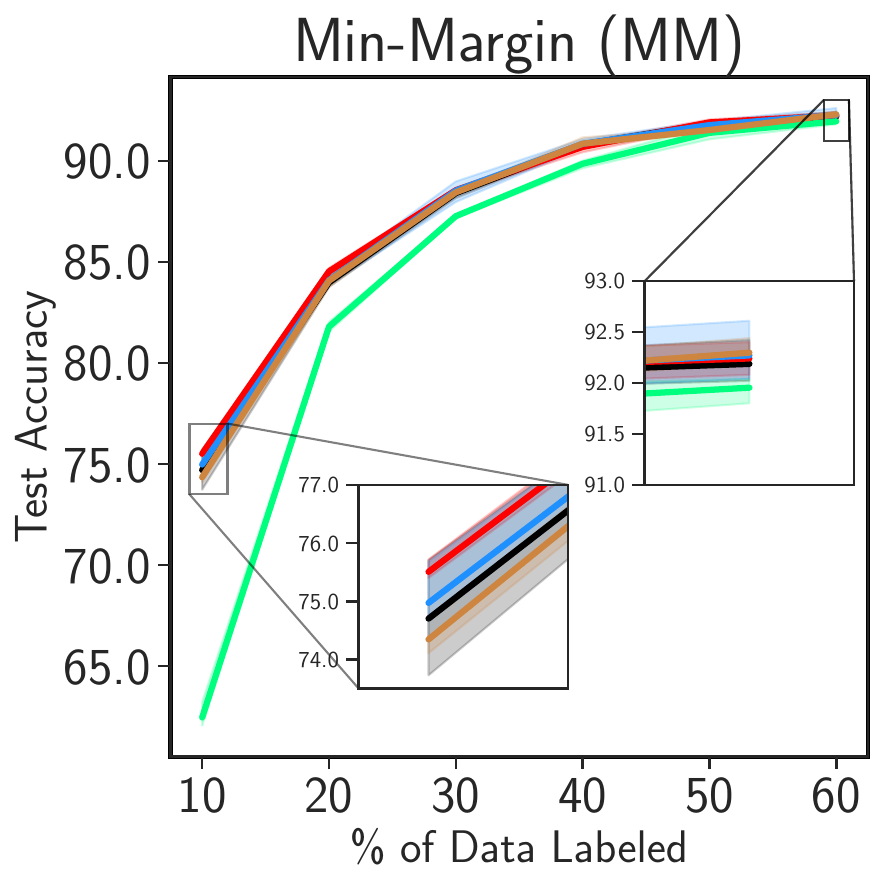}&
\includegraphics[width=0.22\linewidth]{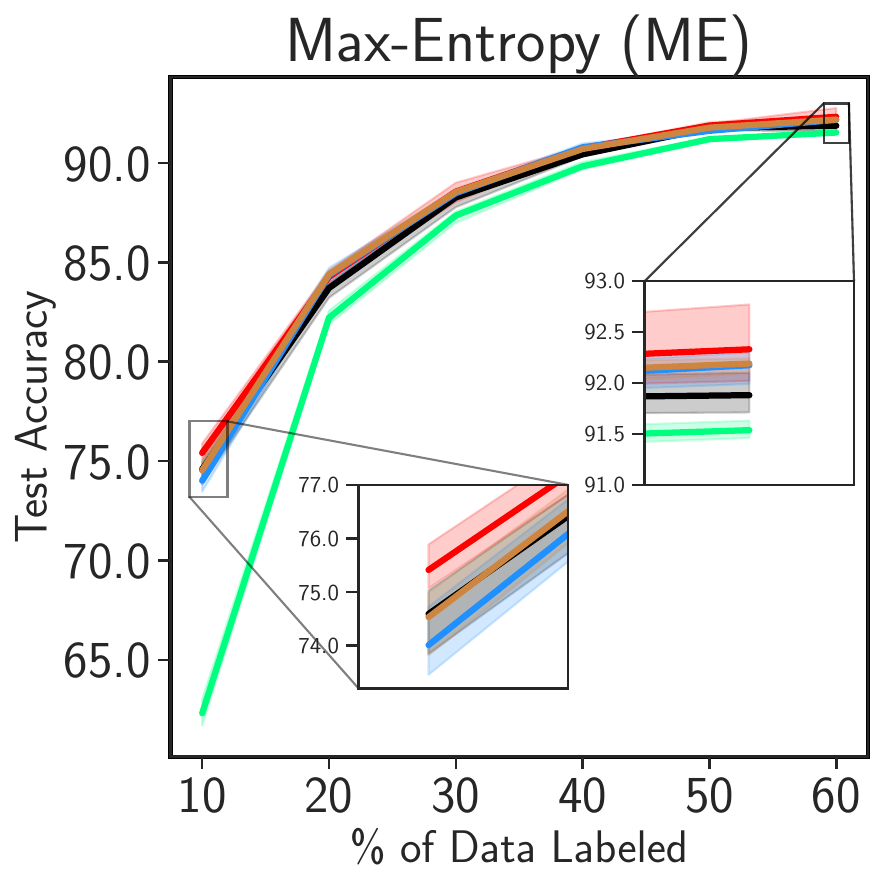}\\

\includegraphics[width=0.22\linewidth]{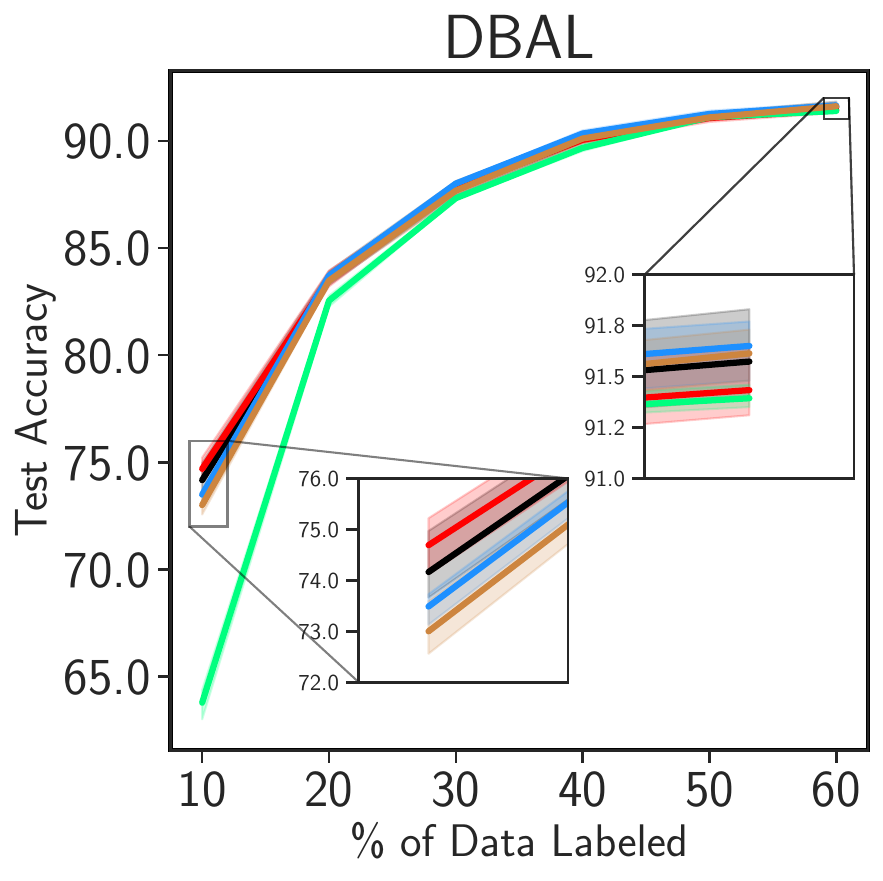}&
\includegraphics[width=0.22\linewidth]{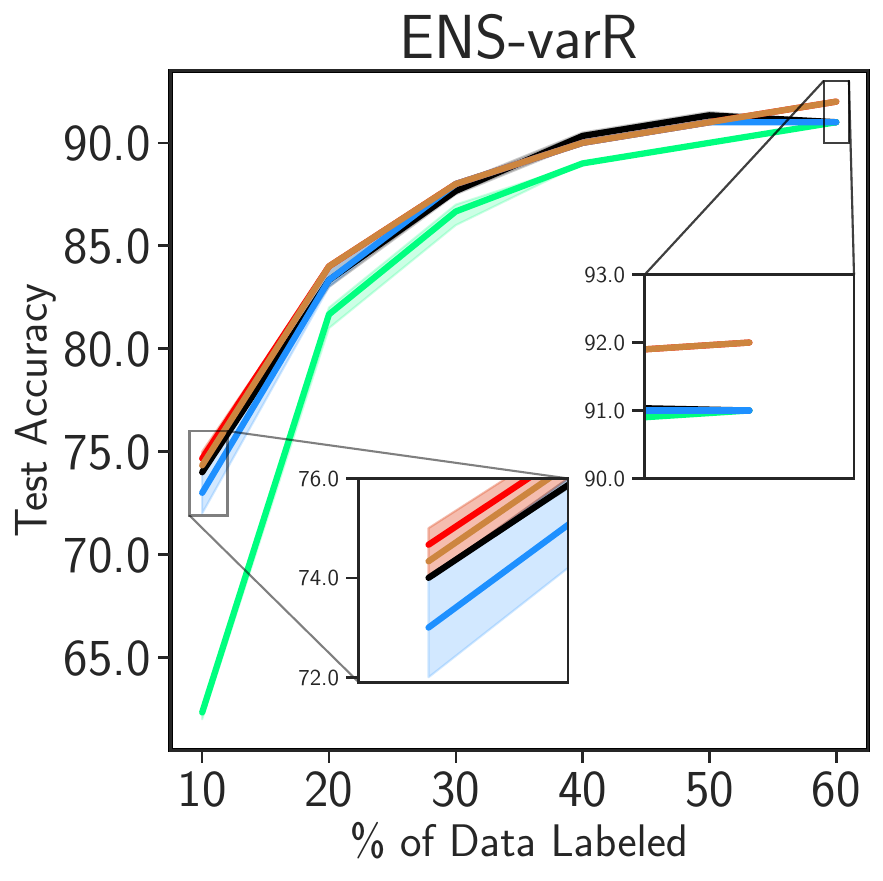}&

\includegraphics[width=0.22\linewidth]{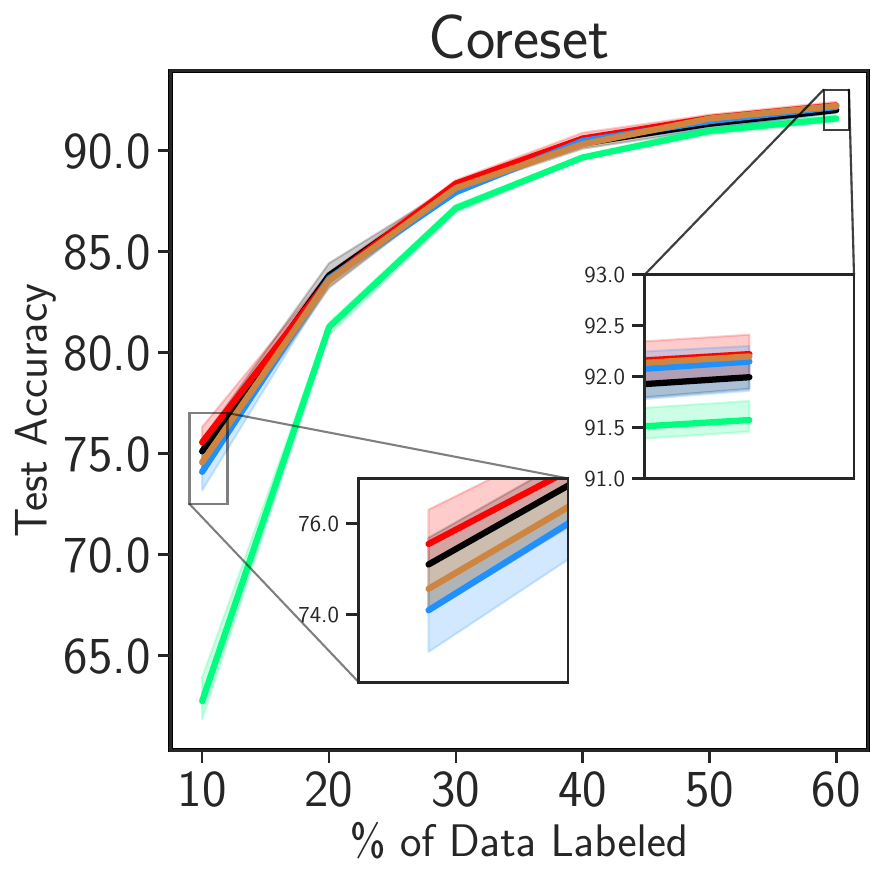}&
\includegraphics[width=0.22\linewidth]{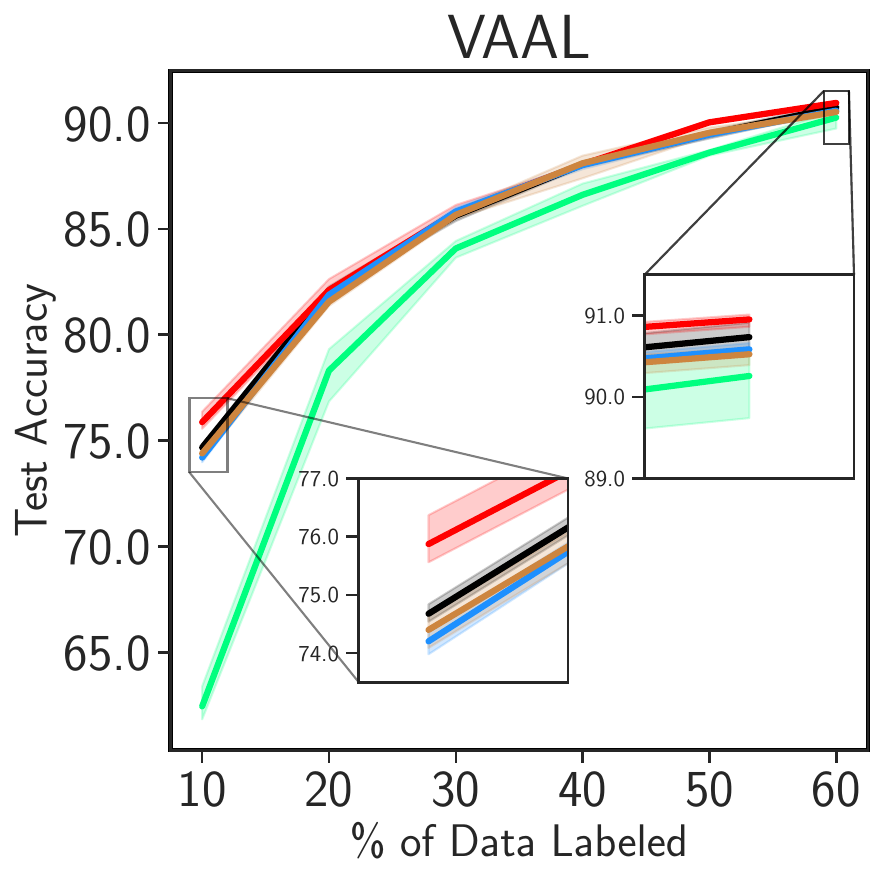}\\

\multicolumn{4}{c}{\includegraphics[width=0.7\linewidth]{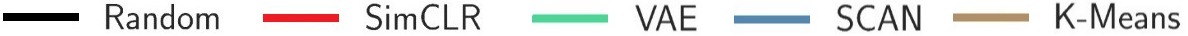}} \\

\end{tabular}
\caption{Performance of each active learning query method with different initial pool sampling strategies on CIFAR-10. There are 8 graphs shown, one for each active querying method as mentioned in the graph titles. Each colored line in the graph corresponds to an initial pool sampling method, as shown in the legend.}
\vspace{-5pt}
\label{fig:cifar10_graphs}
\end{figure}

\section{Experimental Results}

In this section, we first document the modifications to the original experimental protocol. Then, we present the experimental results on MNIST, CIFAR-10, CIFAR-100 and Tiny ImageNet datasets. Then, we discuss our experimental findings and evaluate the extent to which intelligently sampled initial labeled pools help boost AL performance\footnote{Our code is available at \url{https://github.com/acl21/init-pools-dal}}. Finally, we provide more training details needed for reproducing our results. 

\subsection{Modifications to the Original Proposal}

\vspace{3mm} 
{\textbf{Switching to Better Performing Tasks:} } We initially proposed to use image rotation prediction (\cite{rotation_Gidaris2018}) and image inpainting (\cite{inpainting_Pathak2016}) as the self-supervision tasks. To make the experimental setting strong, we switched to a more popular and stronger recent self-supervision method - SimCLR (\cite{simlcr_Chen2020ASF}). In case of unsupervised clustering-based methods, we switched from DeepCluster (\cite{deepcluster_Caron2018}) to a more recent and state-of-the-art clustering method - SCAN (\cite{scan_vangansbeke2020}), which builds on top of SimCLR as a pretext task.

\vspace{3mm}
\noindent
{\textbf{Adding Passive Learning to Experiments List: }}In addition to the AL query methods, we run passive learning (random sampling) cycles starting with intelligently sampled initial pools. This passive learning configuration is equivalent to placebo control group where the intervention wouldn't have direct impact on the final outcome, since we sample data randomly each episode. This is done to check if the performance gains on active learning due to an intelligently labeled initial pool are coincidental.

\vspace{3mm}         
\noindent
{\textbf{Miscellaneous Changes: }}Due to time and computational constraints, we repeated each experiment 3 times instead of the proposed 5 repetitions. We report results on seven out of ten proposed AL query methods, excluding ensemble variants of Least Confidence (ENS-LC), Max-Entropy (ENS-ME) and Min-Margin (ENS-MM). For the same reason, we could not perform the additional binning experiments described in Section \ref{sec:additional-exp}.  We also do not report results of VAAL query method on Tiny ImageNet since one AL cycle with 5 episodes (1 run of experiment) executed for over 100 hours on a single GeForce GTX 1080 Ti GPU. On MNIST, we run 10 episodes of AL starting with 60 instances (0.1\%) in the initial pool and set the AL budget to 60 as well.

\vspace{2.5mm} 
Using grid search, we obtained hyperparameters which are better than the ones mentioned in the proposal. We report the final hyperparameter choices in section \ref{supp}.
 
\begin{figure}
\begin{tabular}{cccc}
\includegraphics[width=0.22\linewidth]{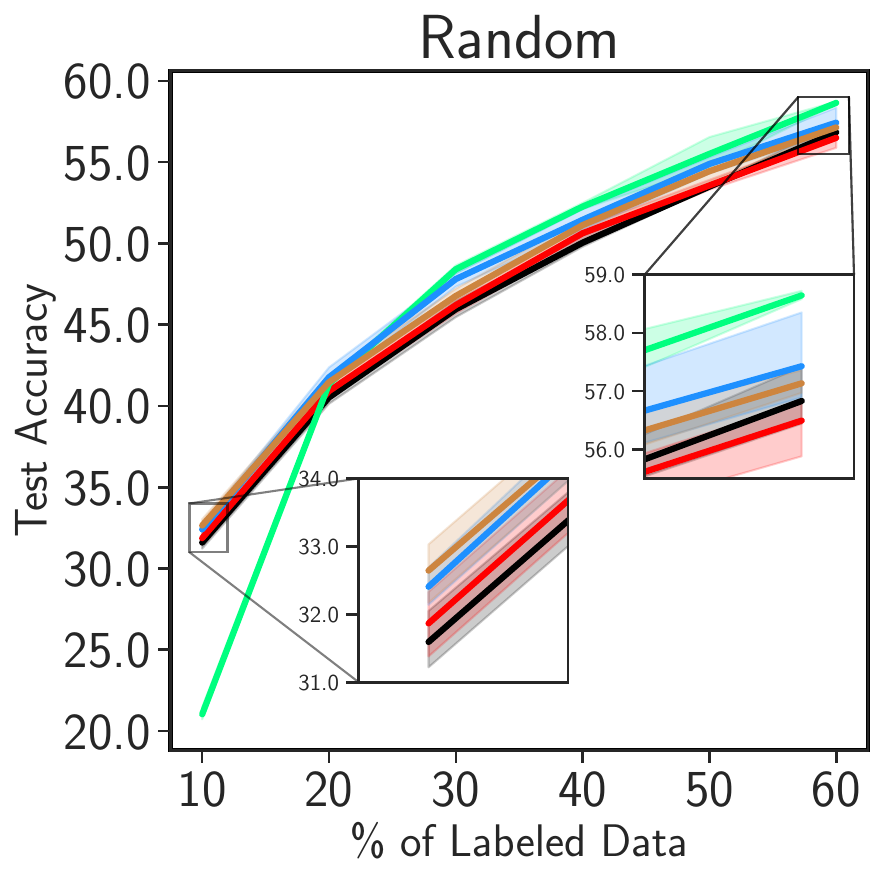}&
\includegraphics[width=0.22\linewidth]{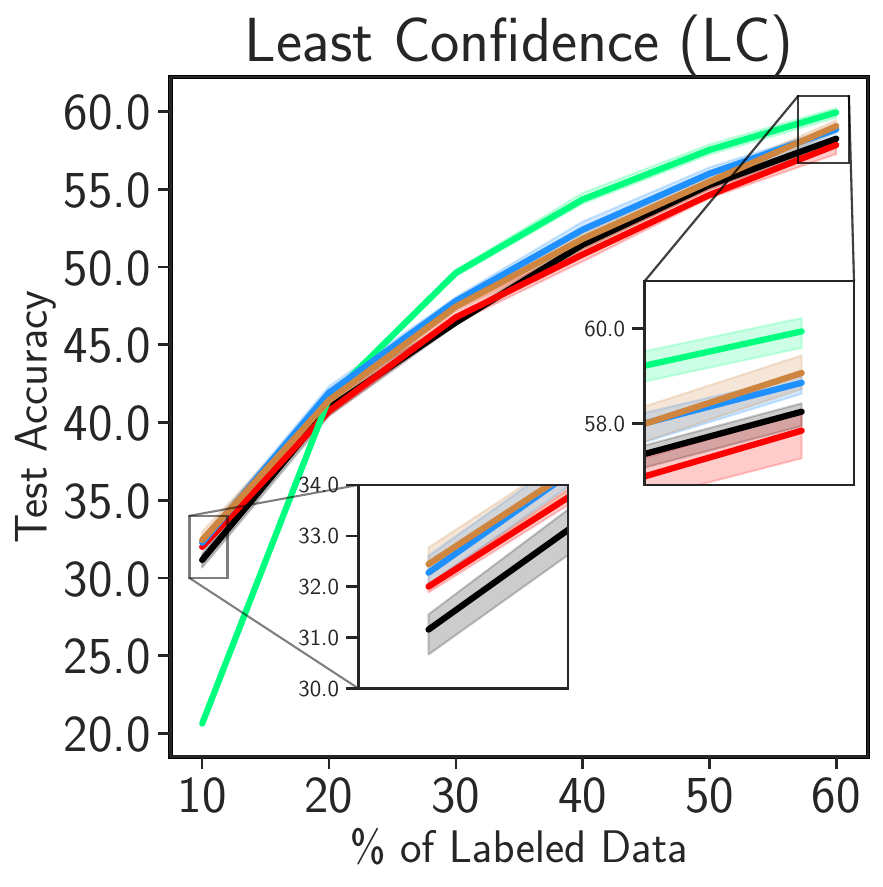}&
\includegraphics[width=0.22\linewidth]{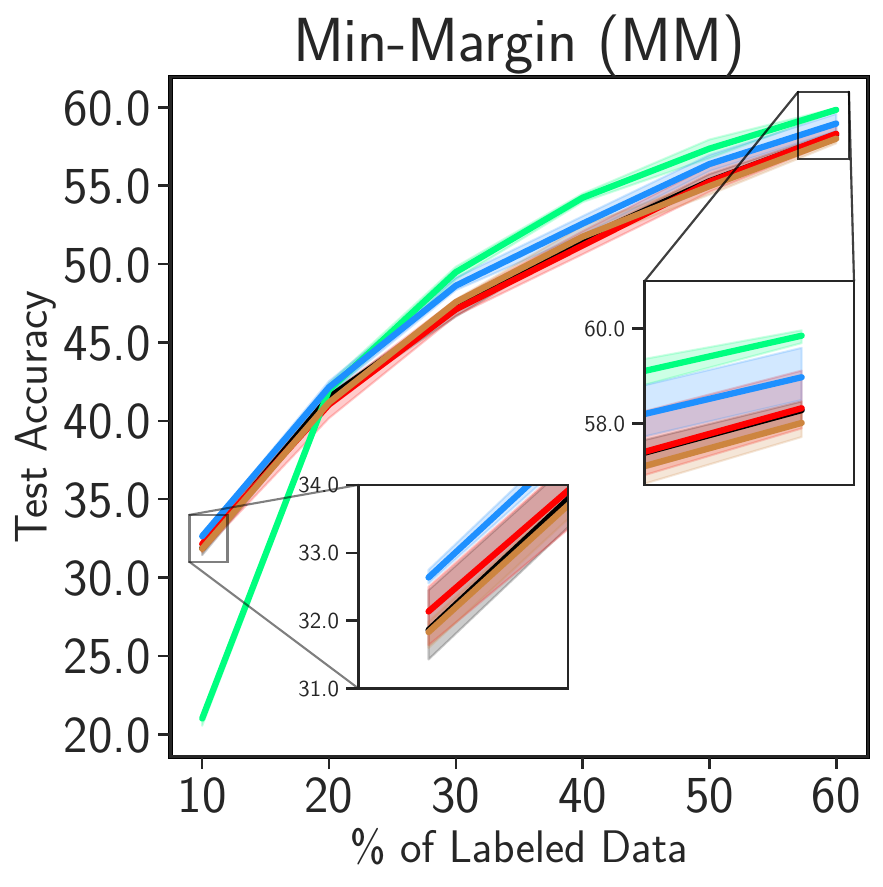}&
\includegraphics[width=0.22\linewidth]{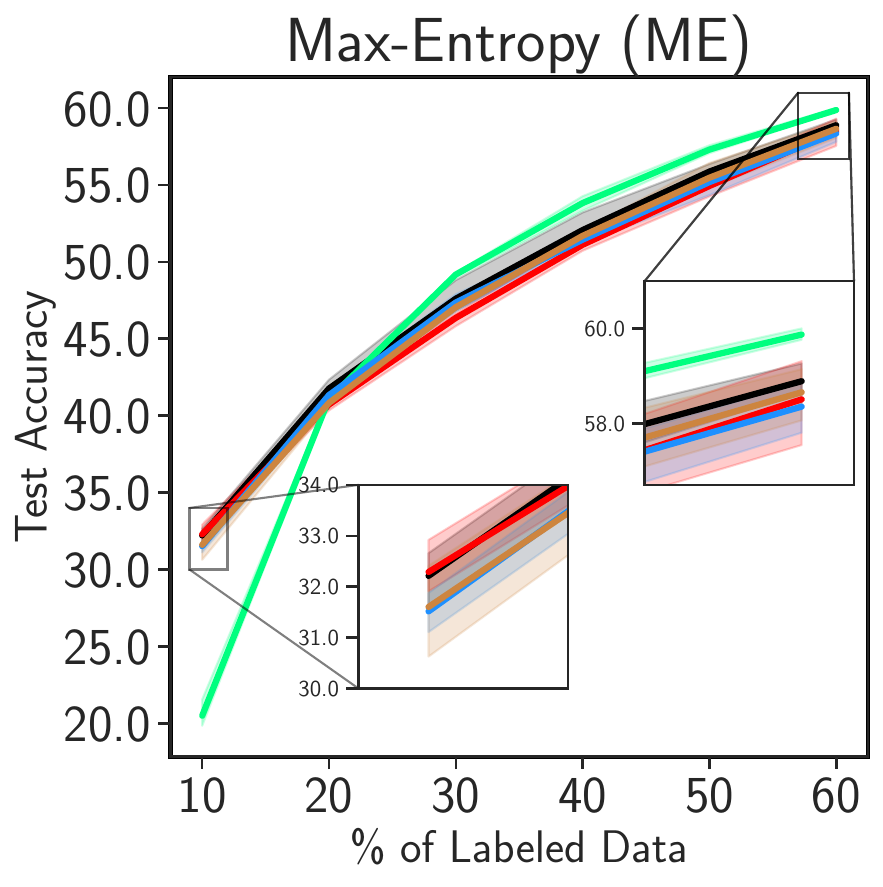}\\

\includegraphics[width=0.22\linewidth]{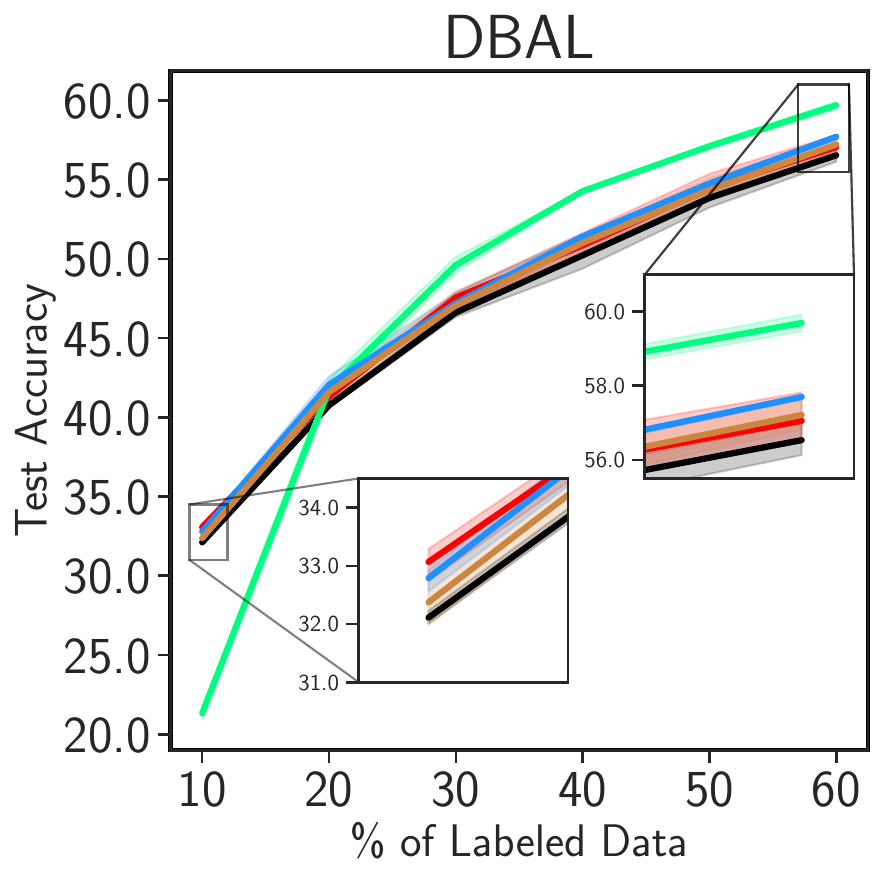}&
\includegraphics[width=0.22\linewidth]{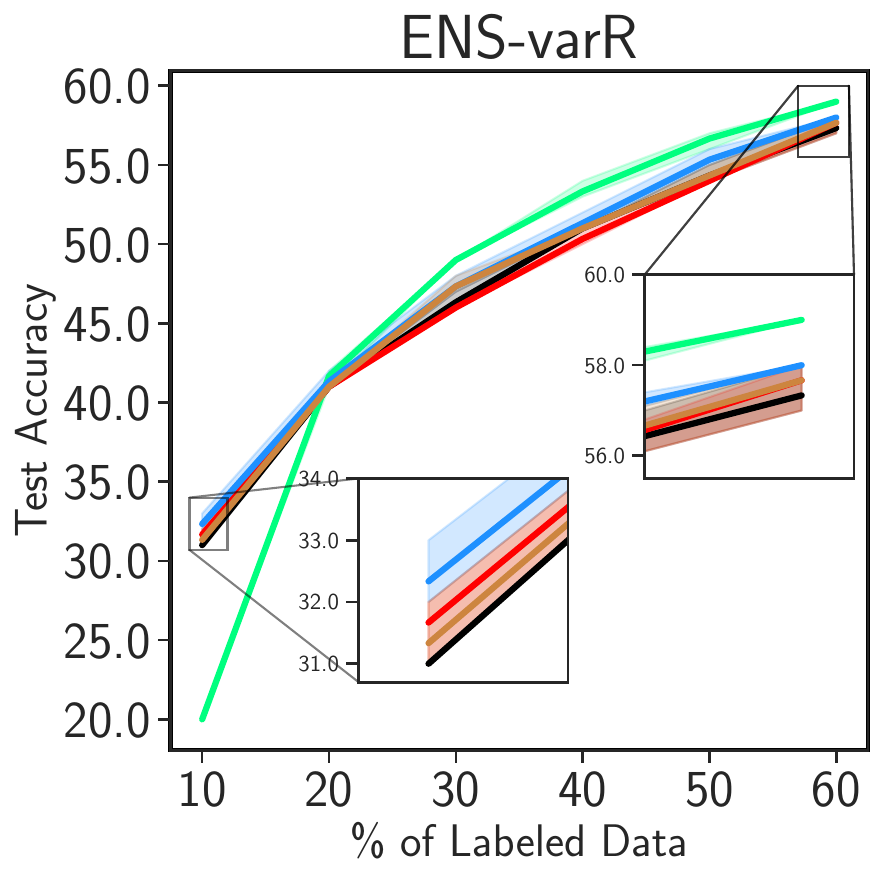}&

\includegraphics[width=0.22\linewidth]{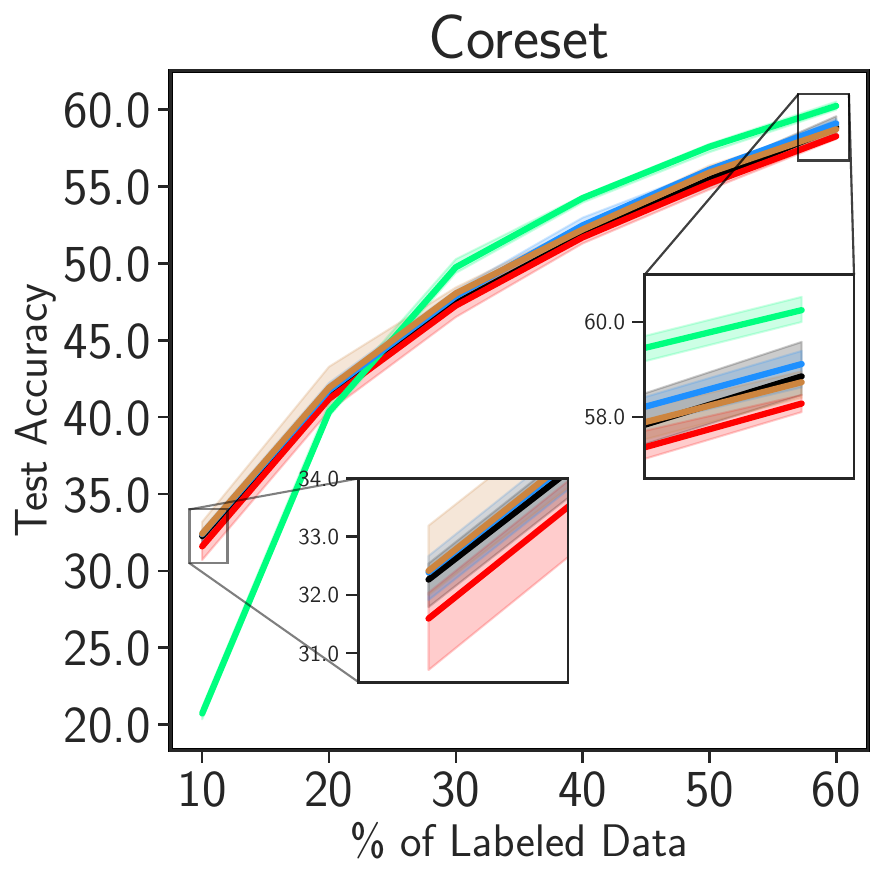}&
\includegraphics[width=0.22\linewidth]{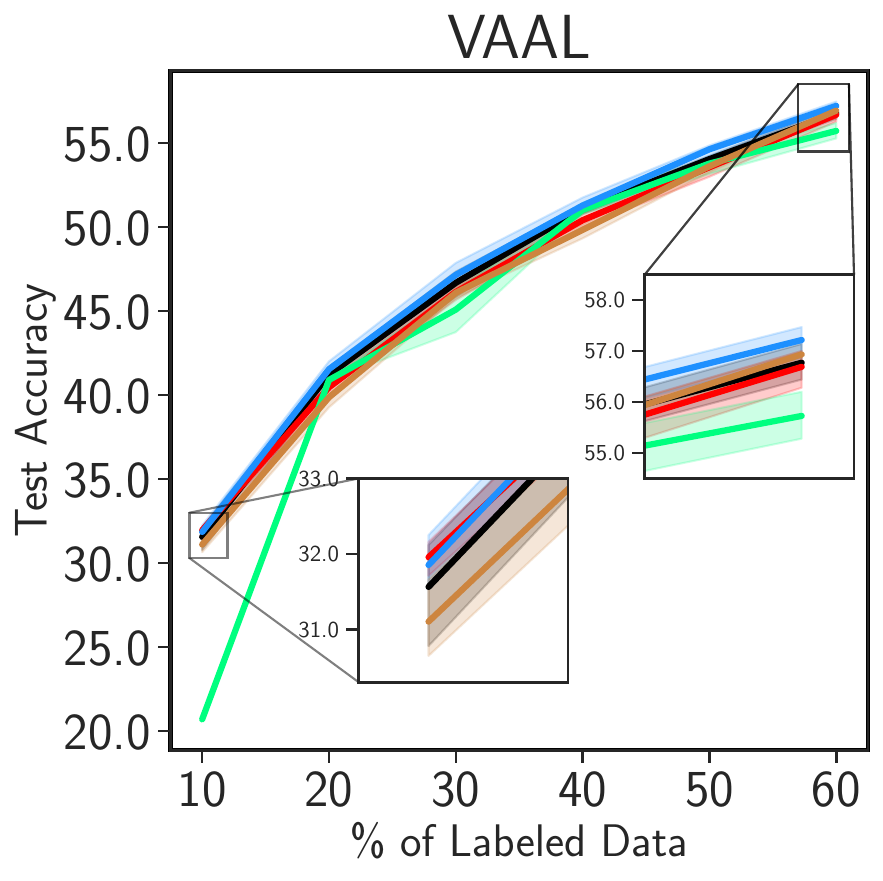}\\

\multicolumn{4}{c}{\includegraphics[width=0.7\linewidth]{figures/legend2.jpg}} \\

\end{tabular}
\caption{AL Performance of each active learning query method with different initial pool sampling strategy on CIFAR-100.}
\vspace{-5pt}
\label{fig:cifar100_graphs}
\end{figure}

\subsection{Initial Pool Sampling Details}\label{sec:init_details}

For completeness, we briefly describe the methods used for our experiments below:

\vspace{3mm}  
\noindent \textbf{SimCLR: }Contrastive learning methods, such as SimCLR (\cite{simlcr_Chen2020ASF}), learn representations by contrasting positive pairs against negative pairs. Positive pairs include input images and their augmented variants. Ideally, a trained SimCLR model should have comparatively low contrastive loss for positive pairs of a given image taken from the unlabeled set. We design our sampling method on this fact. Firstly, we train a ResNet-18 SimCLR model with the recommended augmentations: image horizontal flipping, Gaussian blur, color jitter, and image gray-scale. After training the model, we assign each image in the unlabeled pool a score - model's average contrastive loss between an input image and four of its augmented variants\footnote{MNIST dataset has gray-scale images so we average the contrastive loss over the other three augmentations.}. The higher the average contrastive loss, the \textit{harder} it was for the trained SimCLR model to learn that input, so we sample such images first. 

\vspace{3mm}  
\noindent \textbf{VAE: }We train a vanilla VAE model on the entire training data till convergence. We then sample those data points from the training set whose reconstruction error was high post-training. The higher the reconstruction error, the \textit{harder} it was for the model to learn such images, hence we sample them first. We chose VAE in particular to understand how task complexity (the VAE reconstruction task is simpler than SimCLR's) contributes to initial pool efficiency.

\vspace{3mm}  
\noindent \textbf{SCAN and K-Means: }SCAN (\cite{scan_vangansbeke2020}) is a state-of-the-art clustering method where feature learning and feature clustering are decoupled. SCAN builds on top of features learned by any self-supervision task (in our case, we used SimCLR). At the end of training, the SCAN model assigns a single cluster to each data point. In case of K-Means, we apply K-Means algorithm to SimCLR-learned feature representations to get cluster assignments. Once again, we chose these two clustering methods (one very simple, K-Means, and one more sophisticated, SCAN) to understand the role of model complexity w.r.t. the effectiveness of the initial pool.

\begin{figure}
\begin{tabular}{cccc}
\includegraphics[width=0.22\linewidth]{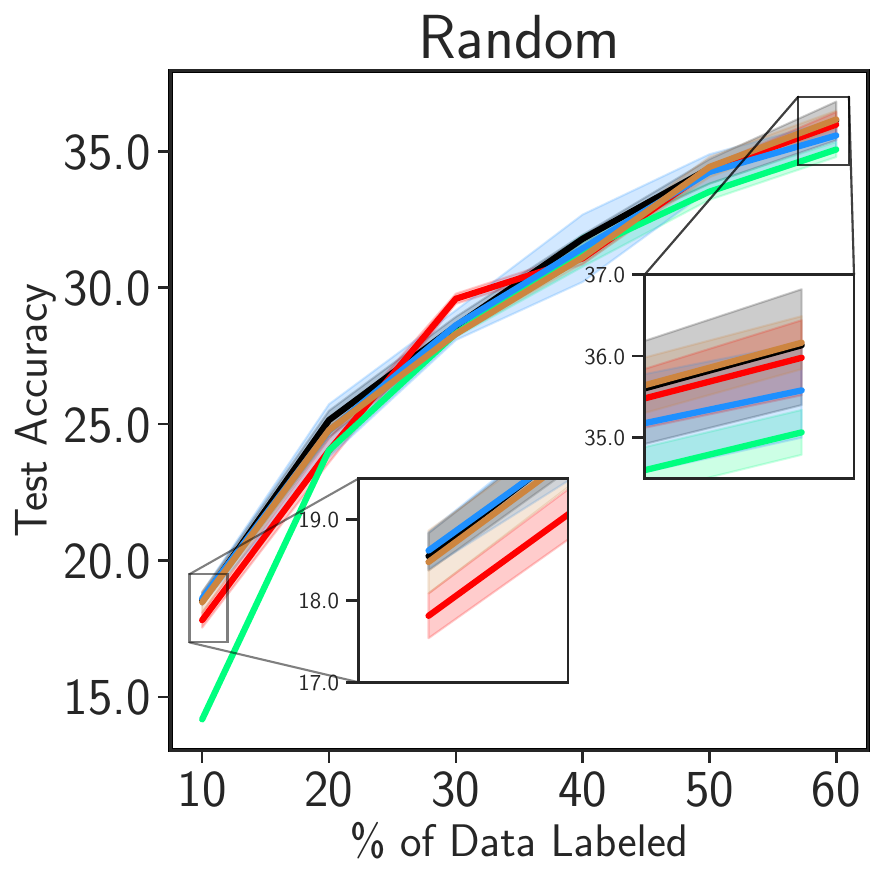}&
\includegraphics[width=0.22\linewidth]{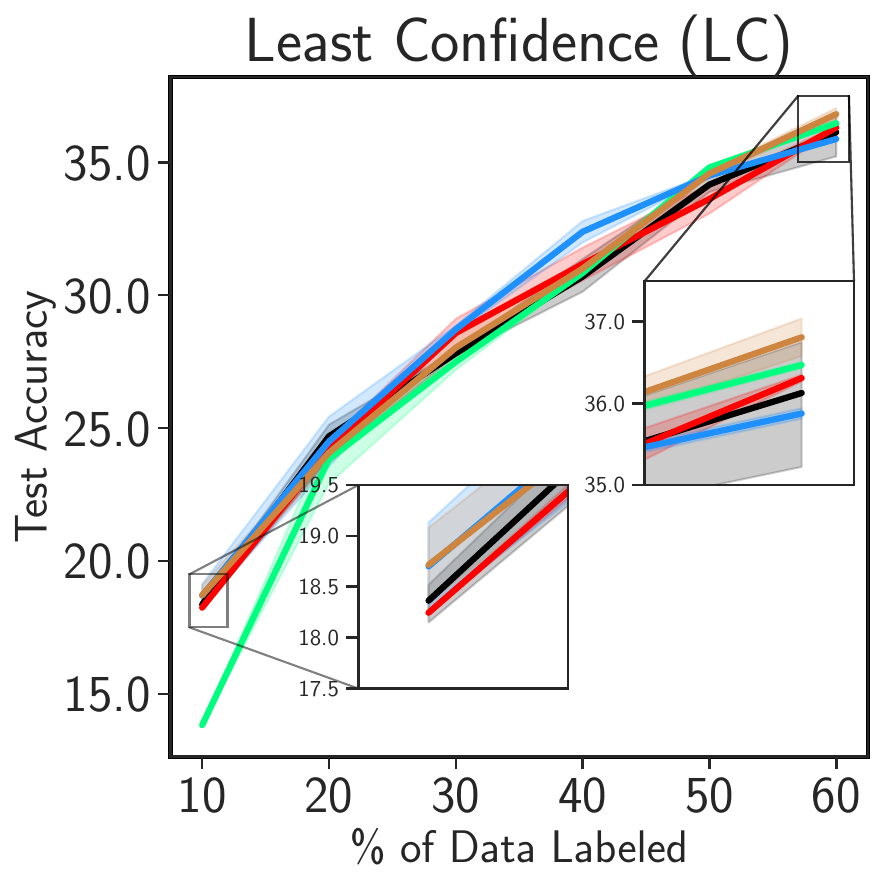}&
\includegraphics[width=0.22\linewidth]{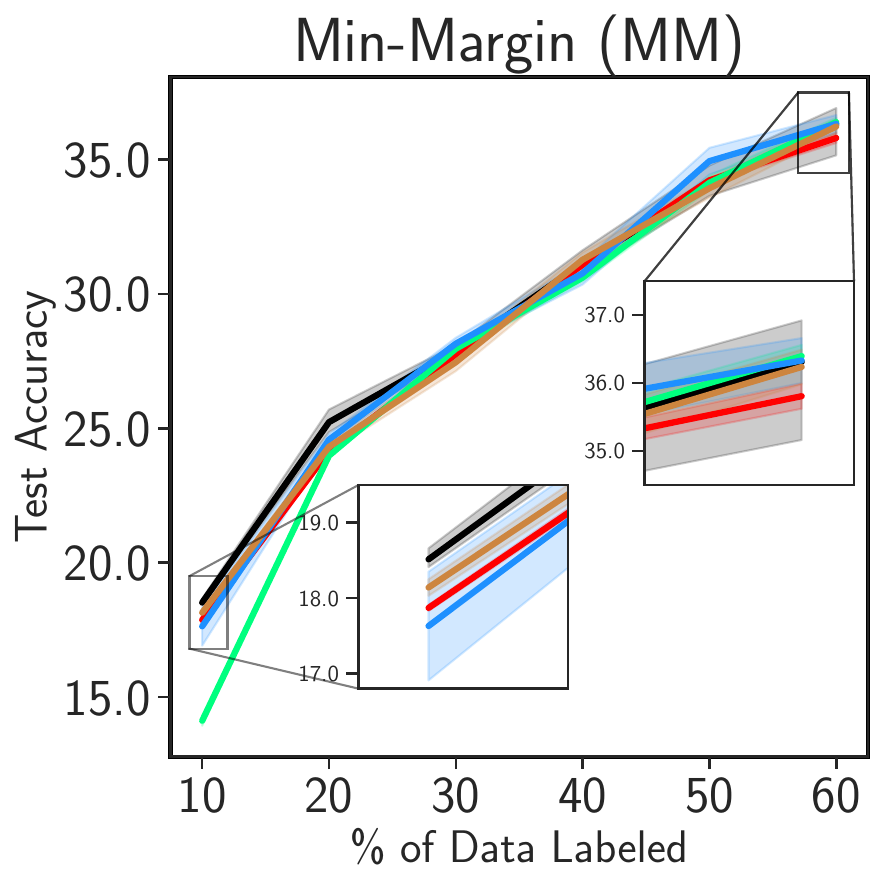}&
\includegraphics[width=0.22\linewidth]{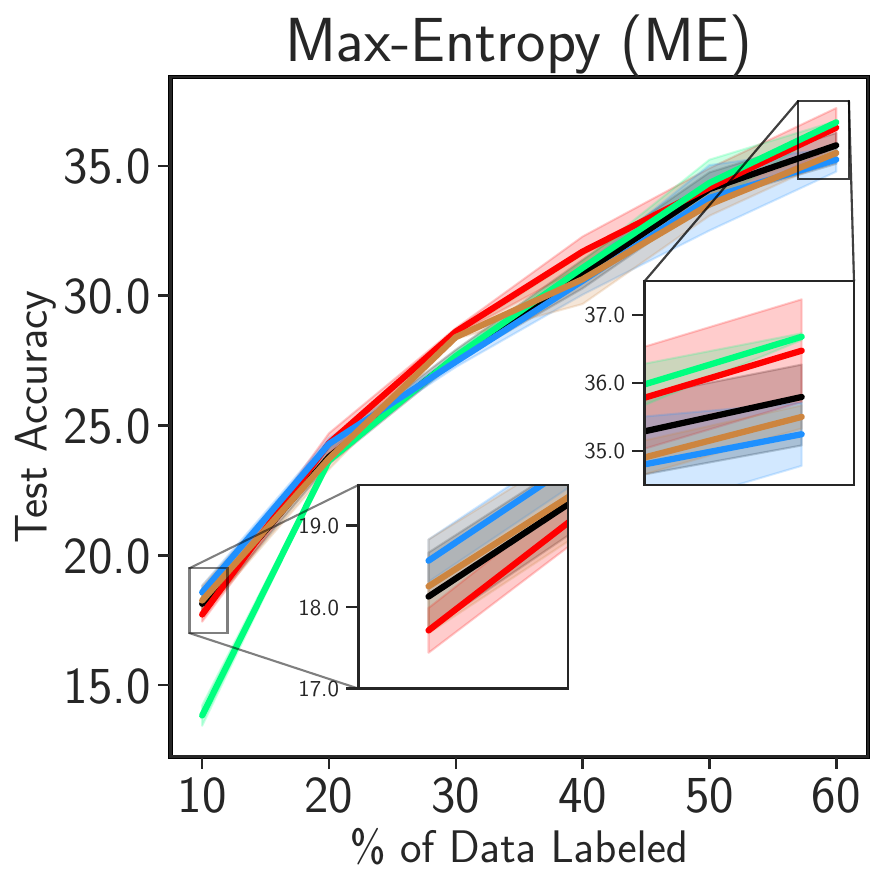}\\

\includegraphics[width=0.22\linewidth]{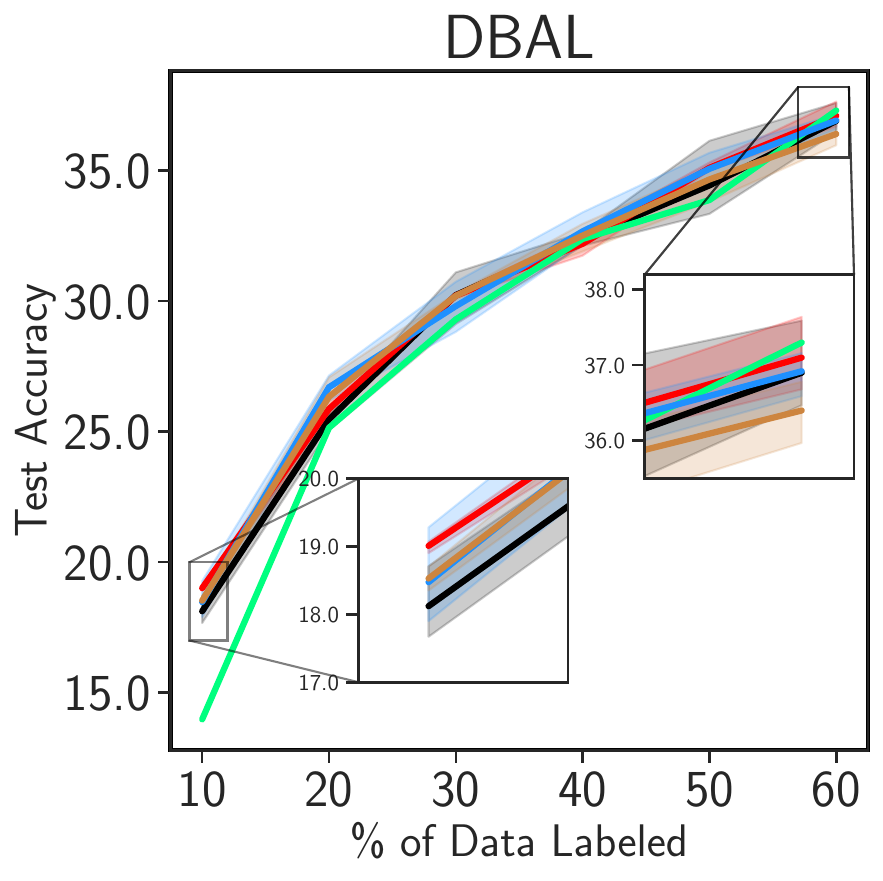}&
\includegraphics[width=0.22\linewidth]{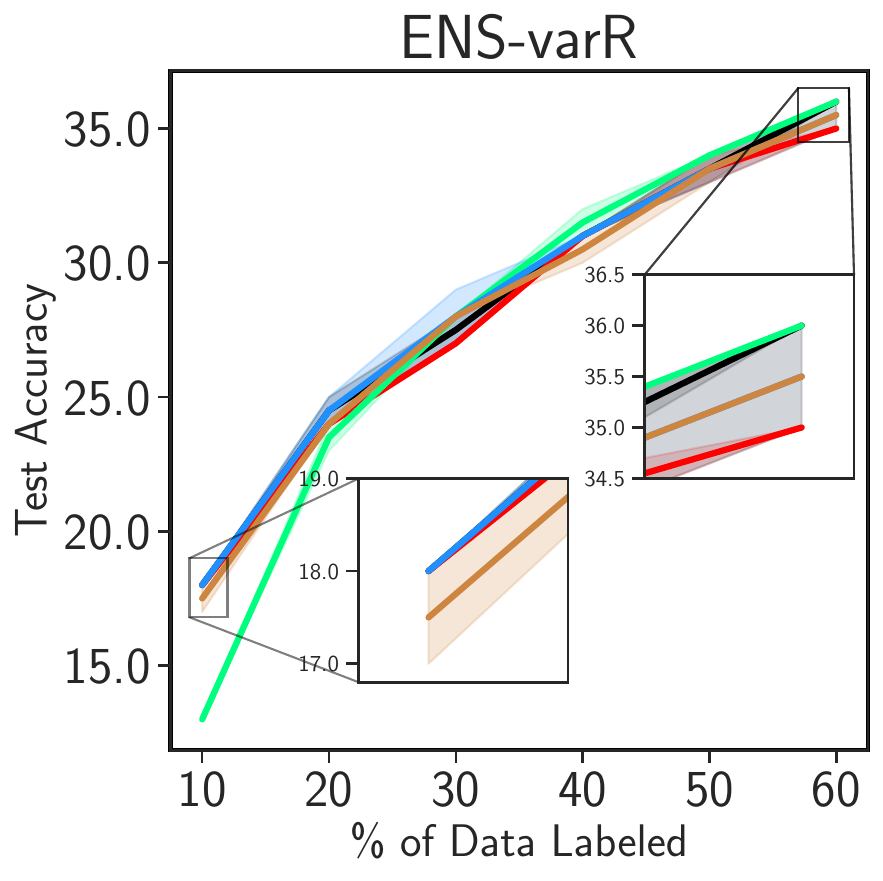}&
\includegraphics[width=0.22\linewidth]{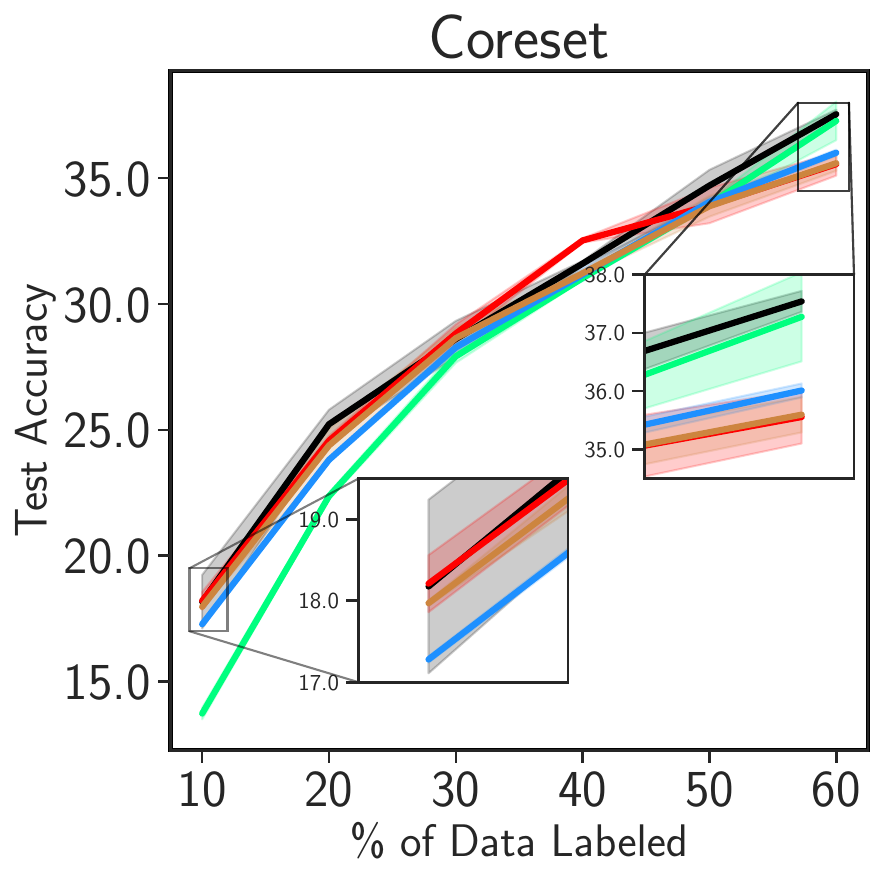}&
\includegraphics[width=0.22\linewidth]{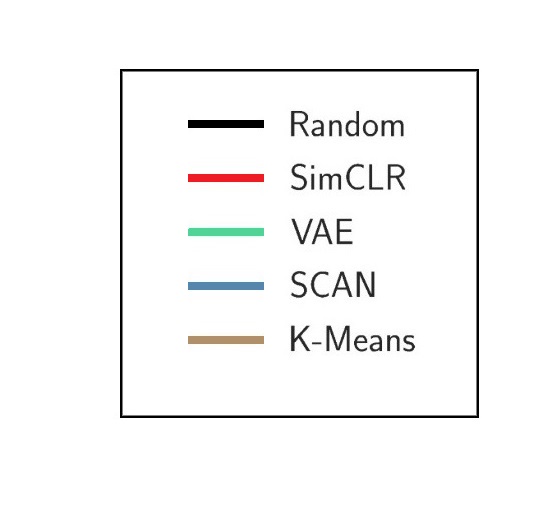}\\

\end{tabular}
\caption{Tiny ImageNet: Our initial pools perform no better than random initial pools across all AL configurations. }
\vspace{-5pt}
\label{fig:tinyimagenet_graphs}
\end{figure}

\subsection{Results}

\subsubsection{Main Experiments}

\begin{figure}
\begin{tabular}{cccc}
\includegraphics[width=0.22\linewidth]{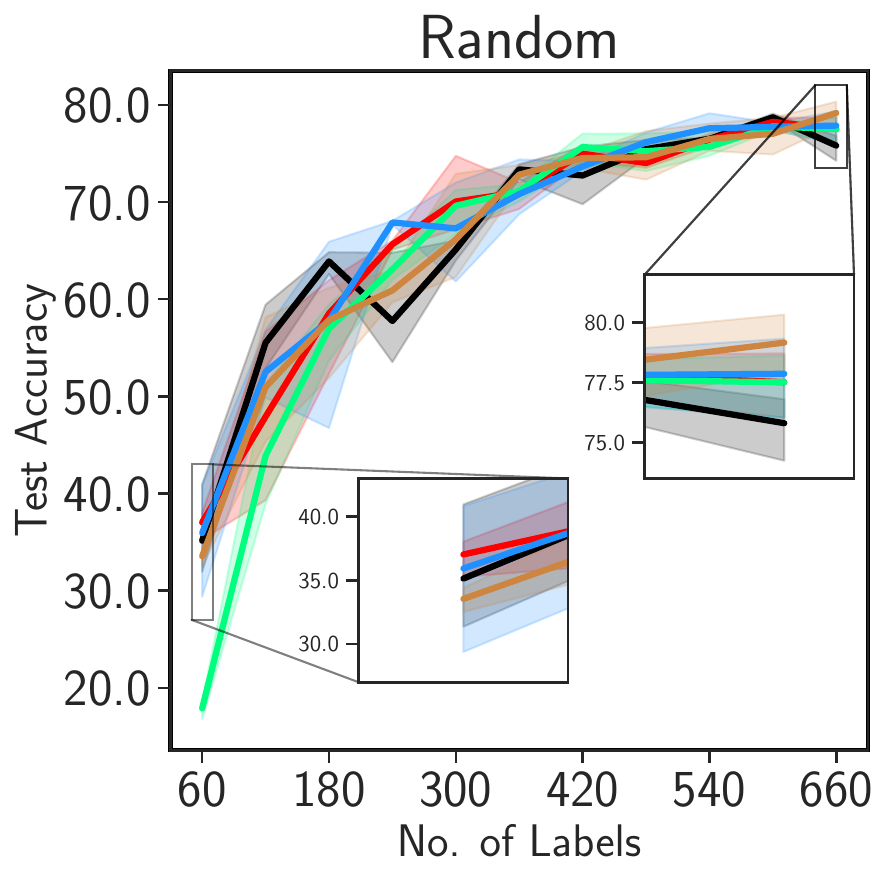}&
\includegraphics[width=0.22\linewidth]{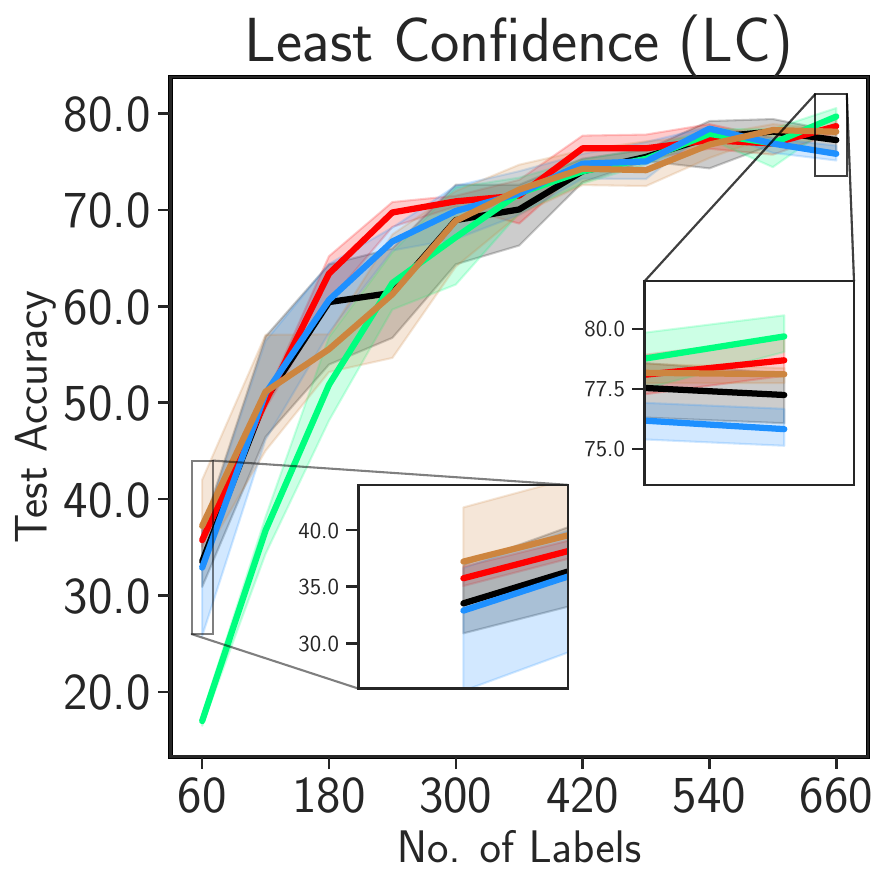}&
\includegraphics[width=0.22\linewidth]{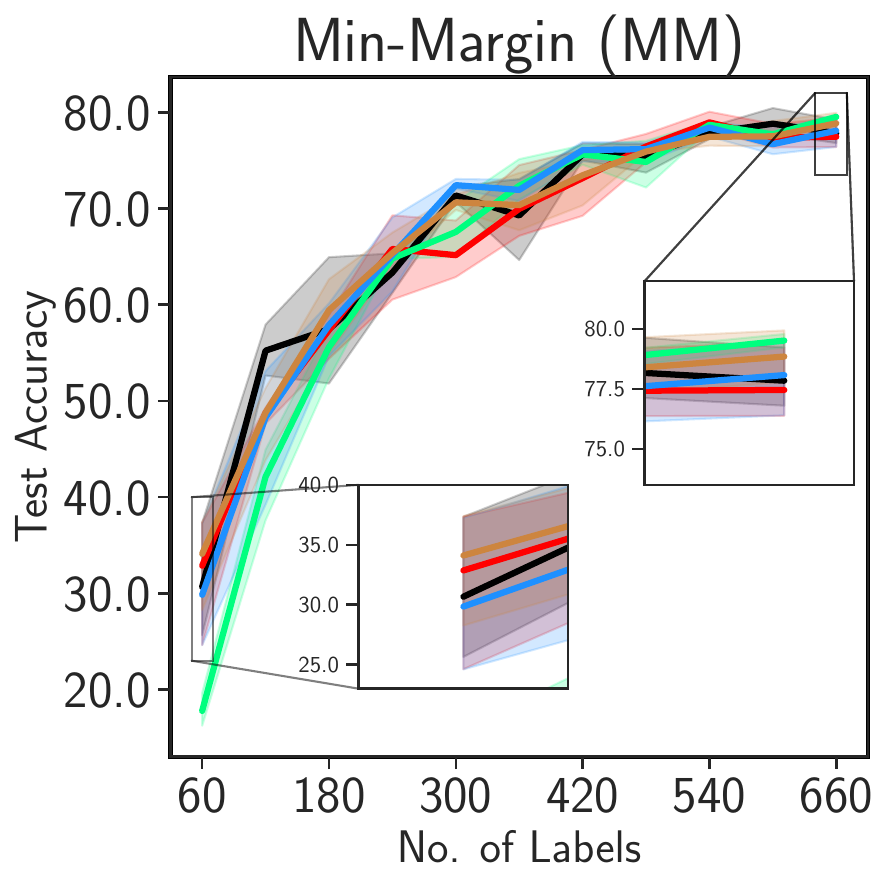}&
\includegraphics[width=0.22\linewidth]{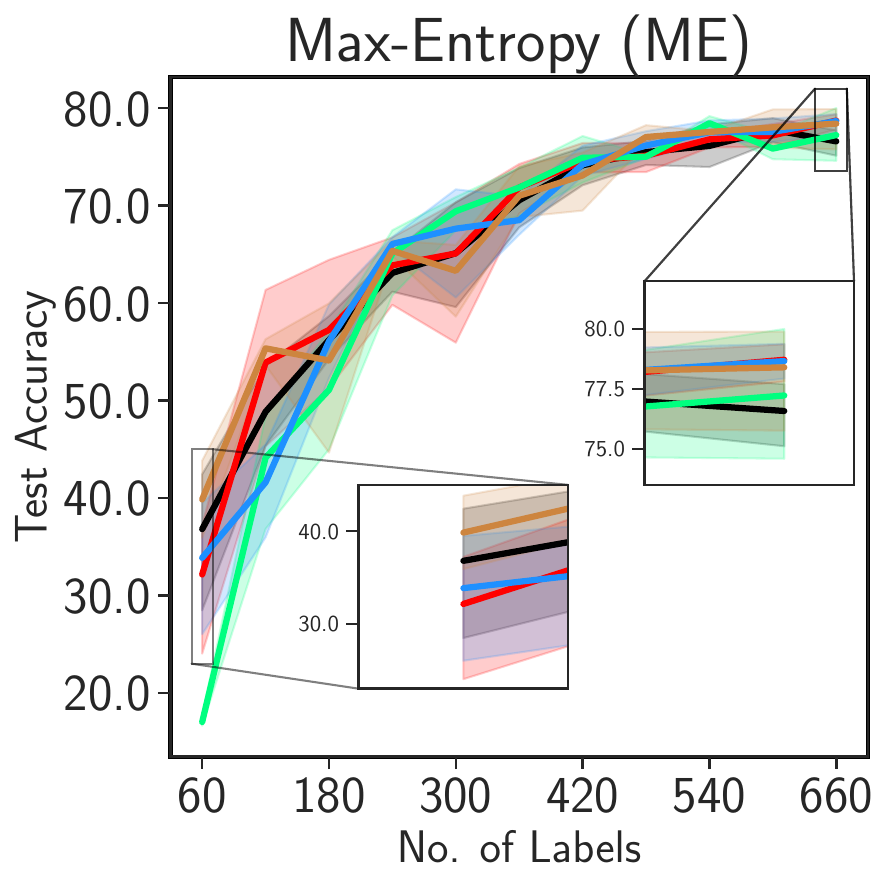}\\

\includegraphics[width=0.22\linewidth]{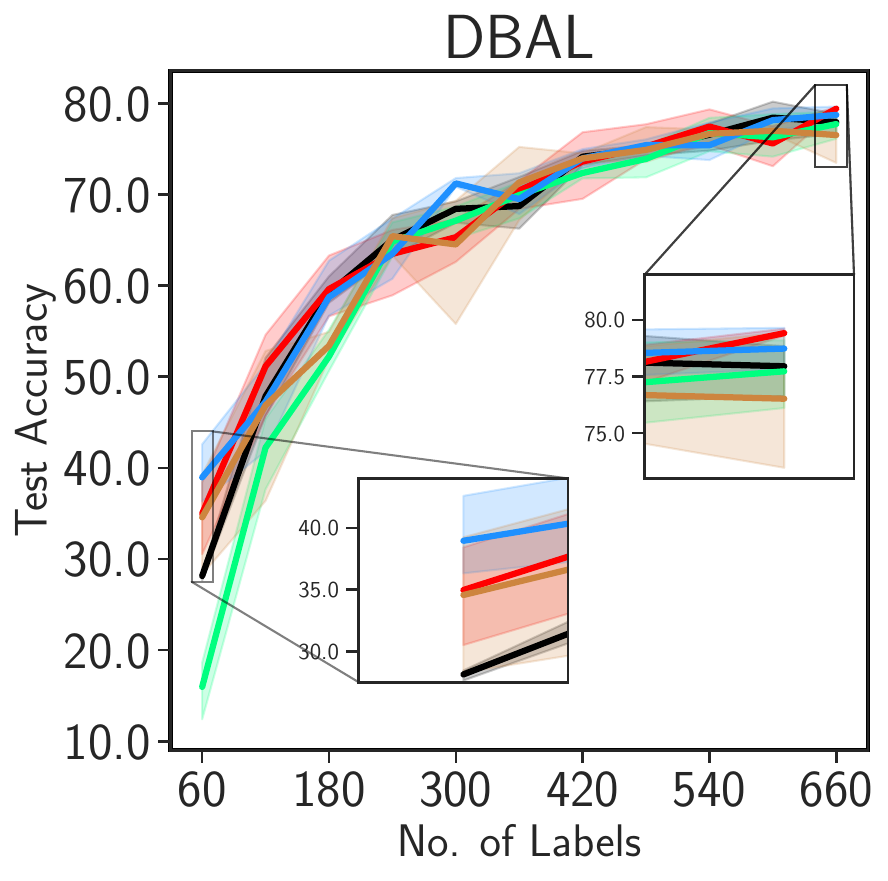}&
\includegraphics[width=0.22\linewidth]{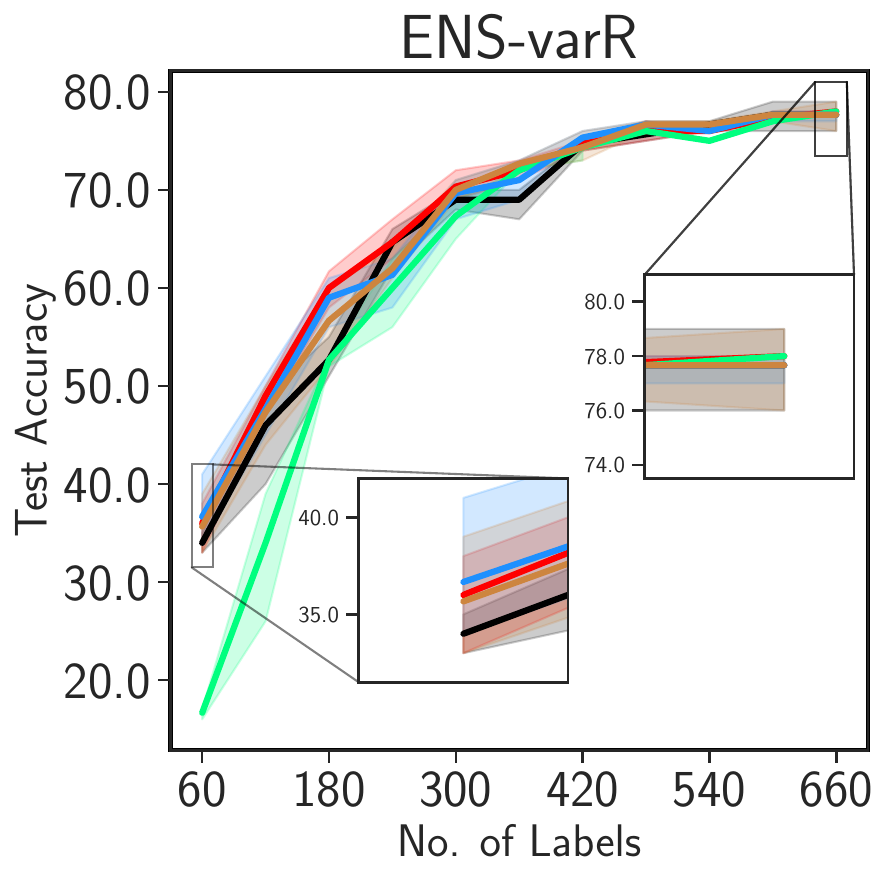}&

\includegraphics[width=0.22\linewidth]{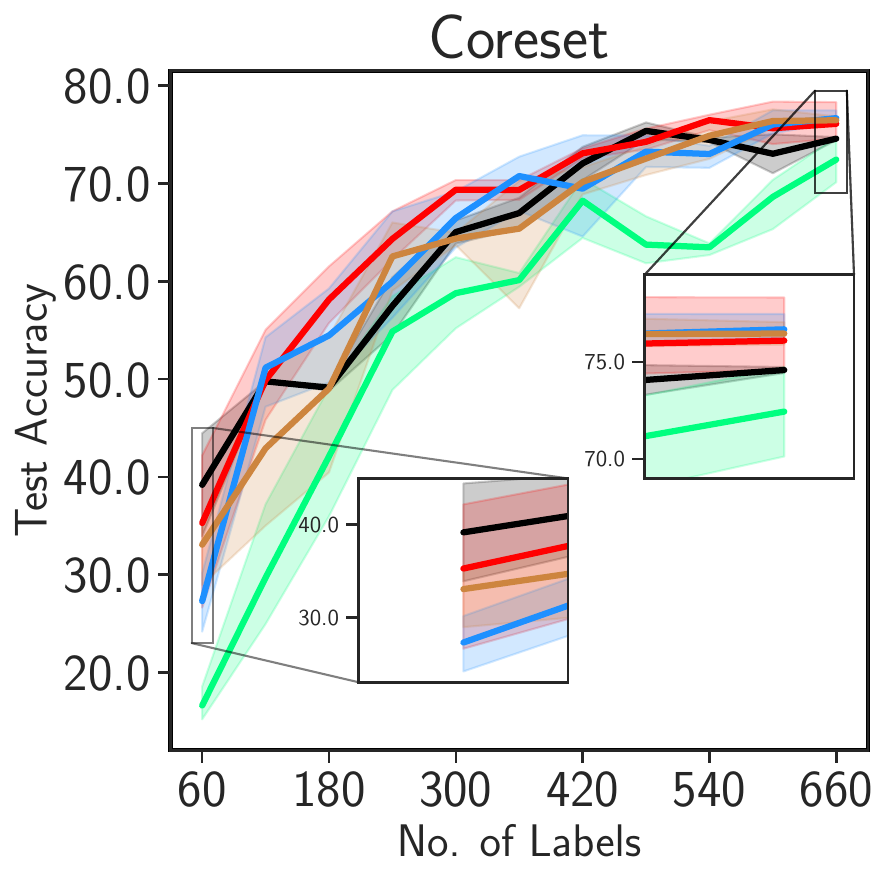}&
\includegraphics[width=0.22\linewidth]{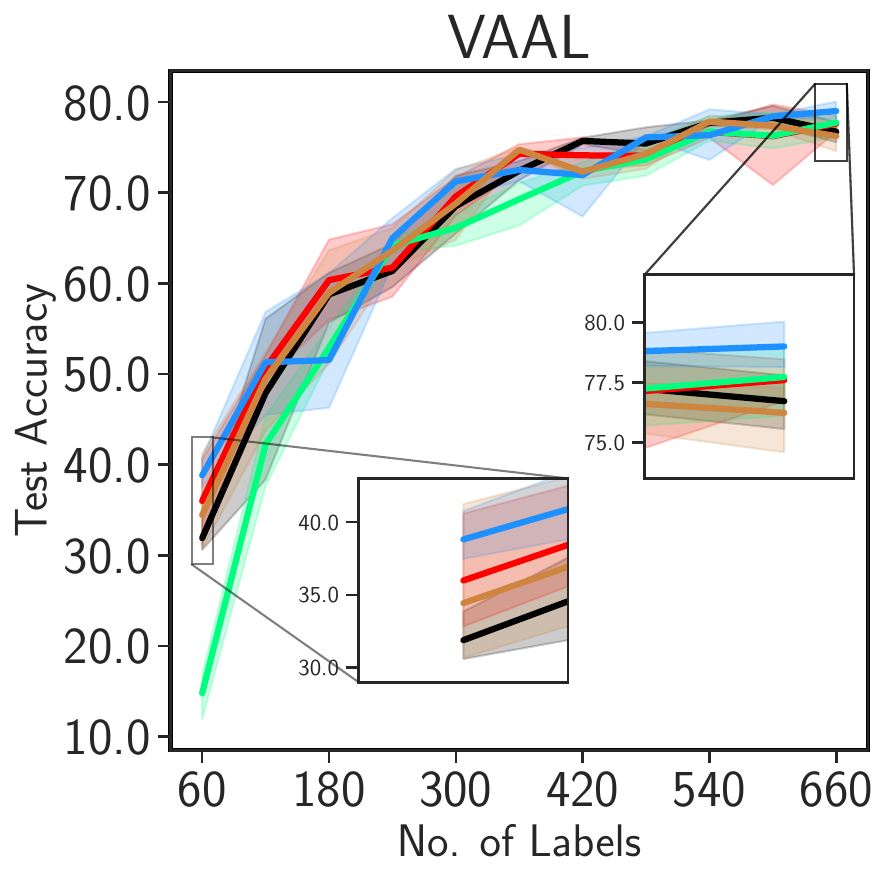}\\

\multicolumn{4}{c}{\includegraphics[width=0.7\linewidth]{figures/legend2.jpg}} \\

\end{tabular}
\caption{MNIST.}
\vspace{-5pt}
\label{fig:mnist_graphs}
\end{figure}

Figures \ref{fig:cifar10_graphs}-\ref{fig:mnist_graphs} depict the main results of our experimental study on finding good initial pools for AL. We first describe how the results are illustrated in Figures \ref{fig:cifar10_graphs}-\ref{fig:mnist_graphs}. We have one figure for each of the following four datasets: CIFAR-10, CIFAR-100, Tiny ImageNet and MNIST. Each plot inside the figures depicts the performance of one AL method with various initial pool techniques. For instance, the plot titled VAAL (second row, right most) in Figure \ref{fig:cifar10_graphs} shows the performance of models trained on data sampled by VAAL's query method in each episode, however initiated with different initial pool strategies (indicated by different colored lines). For example, the red lines show that AL methods were started using the SimCLR-based initial pool strategy. The plots are conventional AL plots where the $x$-axis represents the percentage of labeled data used to train the model, and $y$-axis represents the model's performance on the test set. 

We now briefly analyze the results of each initial pool sampling strategy.

\vspace{3mm} 
\noindent
\textbf{SimCLR: }In Figure \ref{fig:cifar10_graphs} (CIFAR-10), before the first episode, models trained with SimCLR-sampled initial pools show better performance than models trained on a randomly generated initial pool across all eight configurations, including passive learning. However, this performance gain in the beginning of the AL cycles did not contribute to the model in picking better active samples. We can see that models starting with SimCLR-based initial pools performed similar to the models which started with random initial pools at each episode of the AL cycles. Similarly, on CIFAR-100, we see in Figure \ref{fig:cifar100_graphs} that the models starting with SimCLR sampled initial pools perform either same or worse than the models starting on random initial pools at both ends of the AL cycles across all eight configurations. On Tiny ImageNet (Figure \ref{fig:tinyimagenet_graphs}) and MNIST (Figure \ref{fig:mnist_graphs}), we see the same trend as that of CIFAR-100's.

\vspace{3mm} 
\noindent
\textbf{SCAN and K-Means}: Across all datasets, none of the two clustering methods: SCAN and K-Means, show signs of contributing to better model performances compared to random initial pools.

\vspace{3mm} 
\noindent
\textbf{VAE: }Perhaps the most surprising behaviour we noticed among all the methods was how VAE-sampled initial pools worked. Models trained with VAE sampled initial pools consistently underperformed in the first episode of the AL cycles across all four datasets. On CIFAR-10, in the first episode, the average test accuracy difference between models trained with VAE initial pools and other initial pools was 12\%. Similarly, there was a 11\% difference in case of CIFAR-100, 4.5\% in case of Tiny Imagenet and 18\% in case of MNIST. We suspect this is due to the difference in models used for initial pool sampling and active learning. It has been empirically shown that data points actively sampled by one model, say VGG16, do not transfer well to another model, say ResNet-18 (\cite{Lowell2019PracticalOT,Munjal2020TowardsRA}). To allow for smooth transfer of samples, we used the same ResNet18 model for training SimCLR, SCAN and active learning episodes. However, following general trends of use of VAEs, we use a simple VAE model with 4 convolutional layers each in the encoder and the decoder, which may have resulted in this significant difference. 

At the end of all AL cycles, on CIFAR-10, Tiny Imagenet and MNIST, we see that all initial pools converge to largely similar test accuracy, suggesting no significant improvement in AL performance. But we see a different outcome in the case of CIFAR-100 (Figure \ref{fig:cifar100_graphs}). On CIFAR-100, models starting with VAE-sampled initial pools, despite the bad start, ultimately outperform the models starting with the other four initial pools in six out of the seven configurations. In the passive learning configuration (control experiment), we notice that VAE appears to be outperforming others but that happens only in the final episode of the AL cycle, suggesting that this performance gain in six configurations was not a mere coincidence. The models starting with VAE start to outperform their random counterparts right after the second episode (20\%), and we notice the VAE curve starting to diverge from others. However, this behavior was only seen on CIFAR-100. 

To summarize the findings, our proposed methods could not conclusively prove the existence of \textit{good} initial pools that help AL methods in the long run, although the use of VAE-based initial pool strategy showed some interesting trends.

\vspace{3mm} 
\noindent
\textbf{What explains the odd behavior of VAE sampled initial pools on CIFAR-100?} \\
To investigate the reason behind the odd behavior of models started with VAE-based initial pools, we study the class distribution of initial pools obtained by 4 methods - VAE, SCAN, SimCLR and K-Means. To this end, we picked initial pools from the DBAL experiment.

\begin{figure}
\begin{tabular}{cc}
\includegraphics[width=0.48\linewidth]{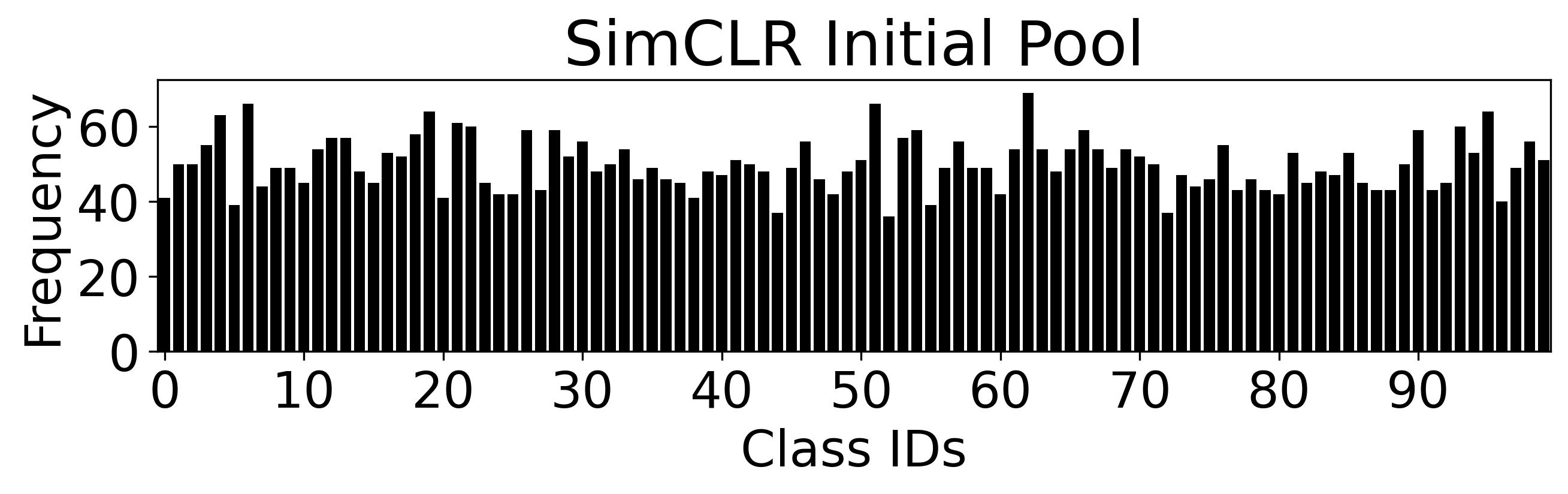}&
\includegraphics[width=0.48\linewidth]{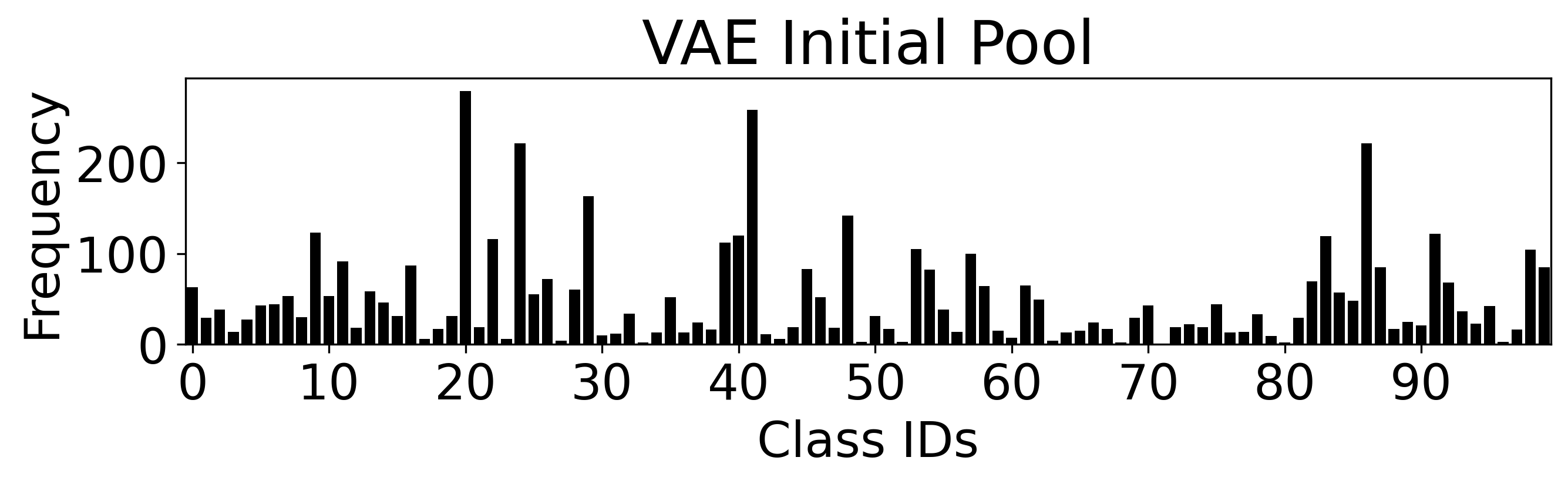}\\
\includegraphics[width=0.48\linewidth]{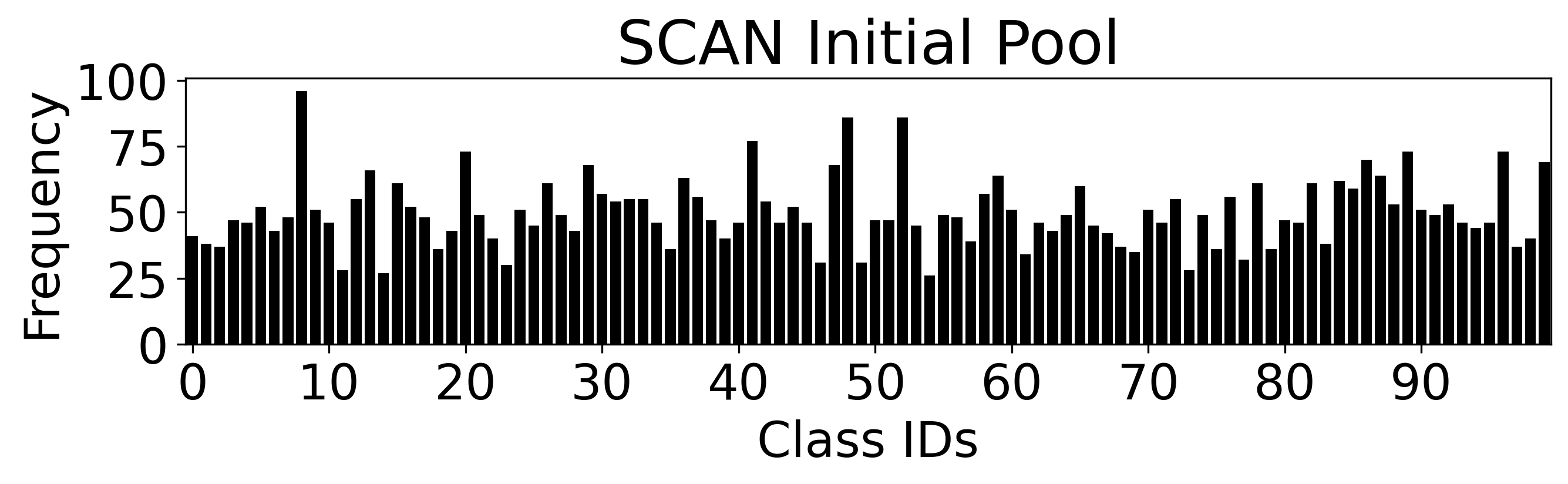}&
\includegraphics[width=0.48\linewidth]{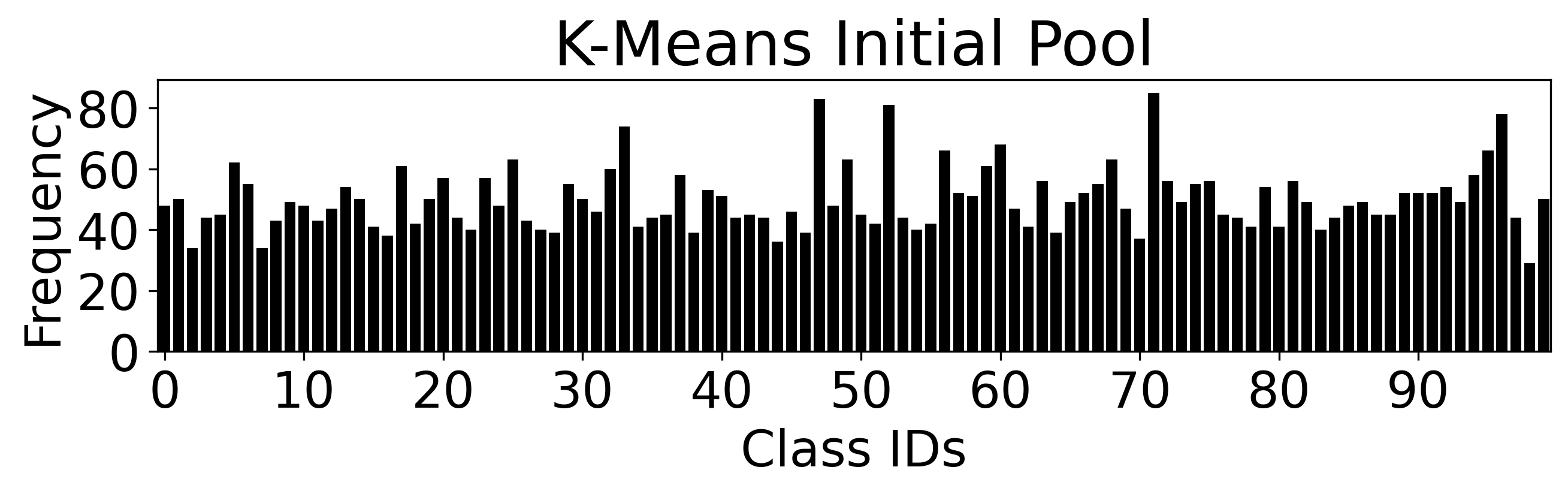}\\

\end{tabular}
\caption{CIFAR-100: Class distribution of initial pools picked by various methods. Note the apparent class imbalance in the initial pool picked by VAE. Is this the reason for the performance gain?}
\vspace{-5pt}
\label{fig:class_frequency_graphs}
\end{figure}

Looking at the class frequencies of all CIFAR-100 initial pools in Figure \ref{fig:class_frequency_graphs}, we notice a clear difference between the VAE-sampled initial pool and the others. VAE-sampled initial pool has more class imbalance and is particularly emphasizing on images from specific classes. To verify if the VAE-based initial pool sampling technique is biased towards difficult classes, we created two sets of CIFAR-100 classes: (1) Top 10 classes sampled by VAE, (2) 10 classes with least per-class test accuracy w.r.t. the model in the final AL episode. We use (2) as a proxy for ``difficult'' classes. We observed that both these sets have 4 classes in common. While this overlap is not high enough to conclude that VAE sampling is biased towards difficult classes, it nevertheless is an interesting future direction to pursue, and merits more study.

\subsubsection{Ablation Experiments}

\begin{figure}[h]
\begin{tabular}{ccc}
\includegraphics[width=0.3\linewidth]{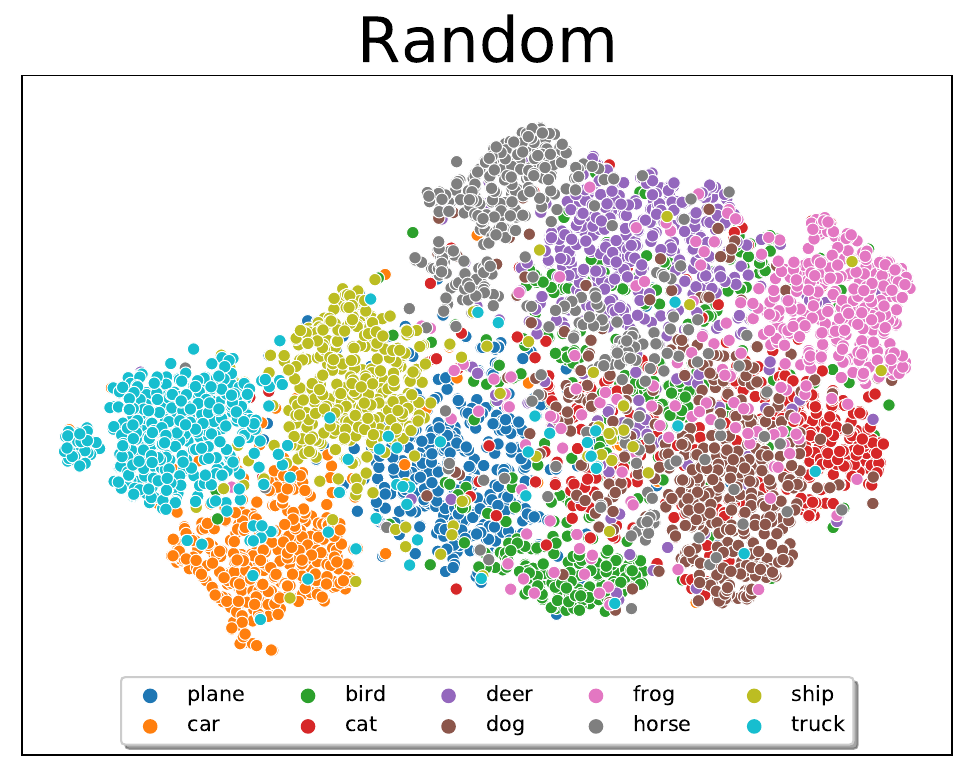} &
\includegraphics[width=0.3\linewidth]{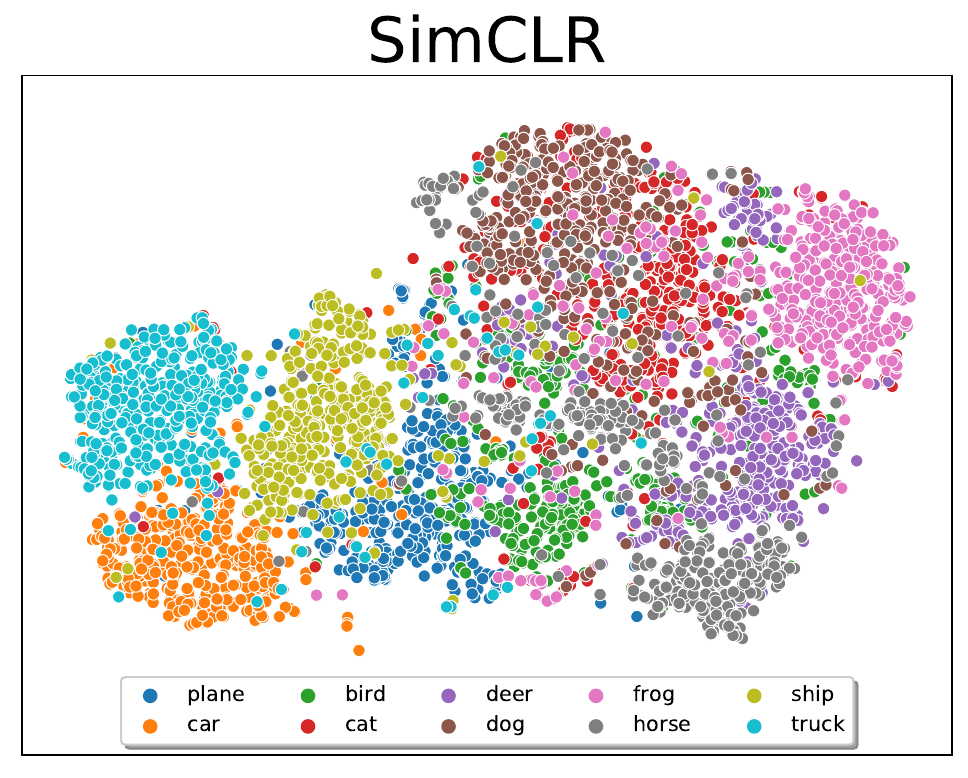} &
\includegraphics[width=0.3\linewidth]{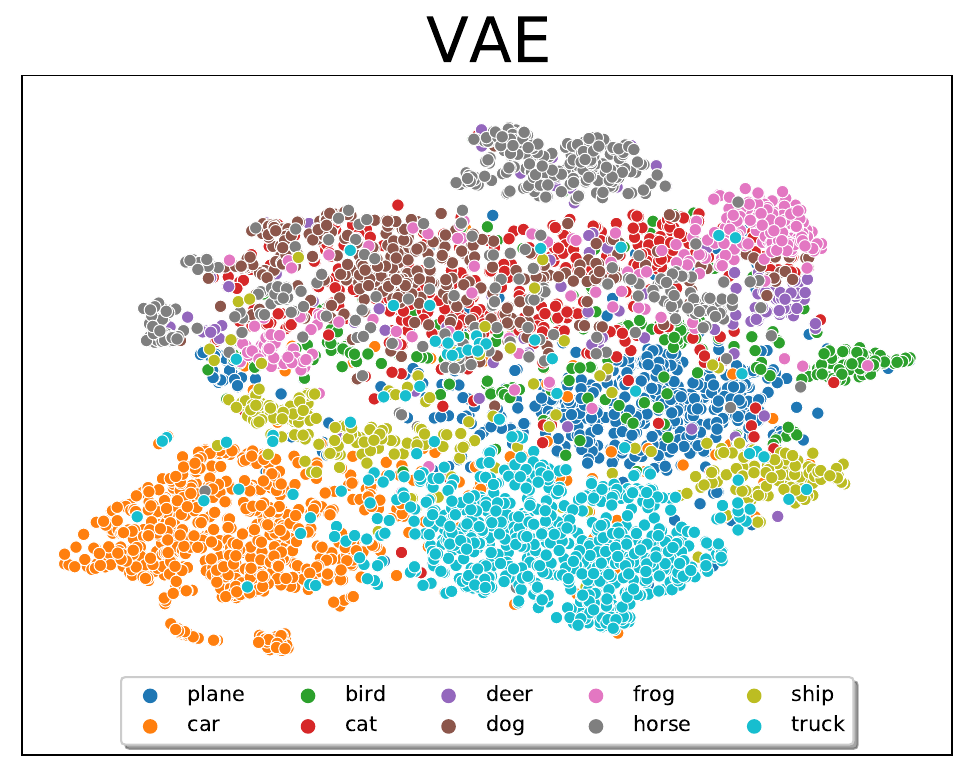} \\

\includegraphics[width=0.3\linewidth]{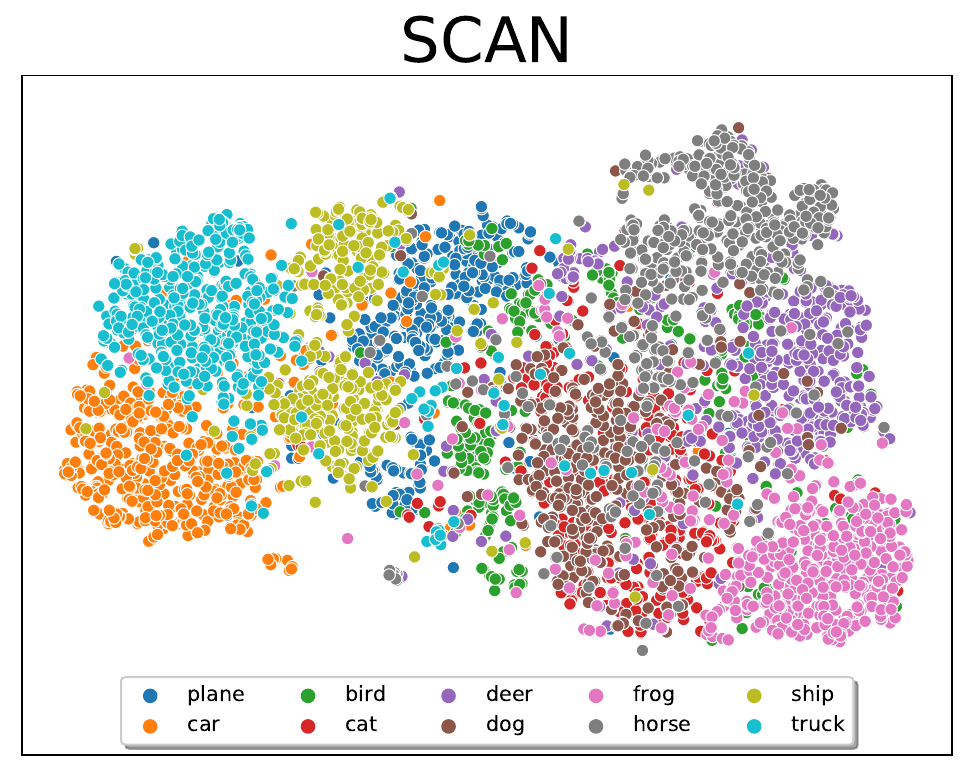} &
\includegraphics[width=0.3\linewidth]{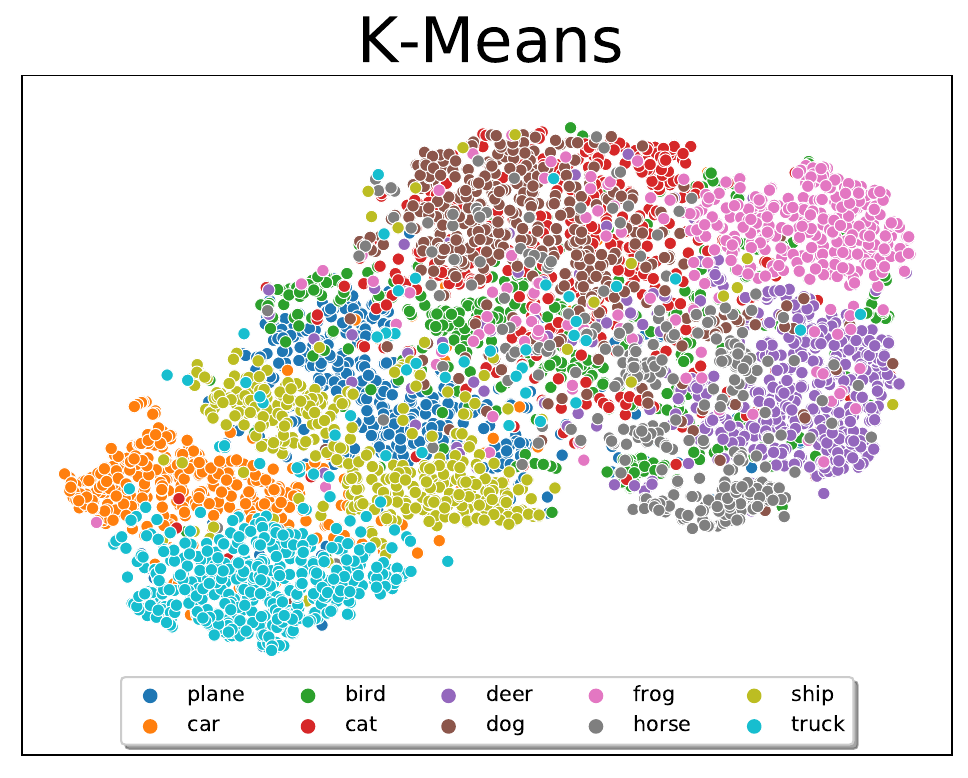} &
\includegraphics[width=0.3\linewidth]{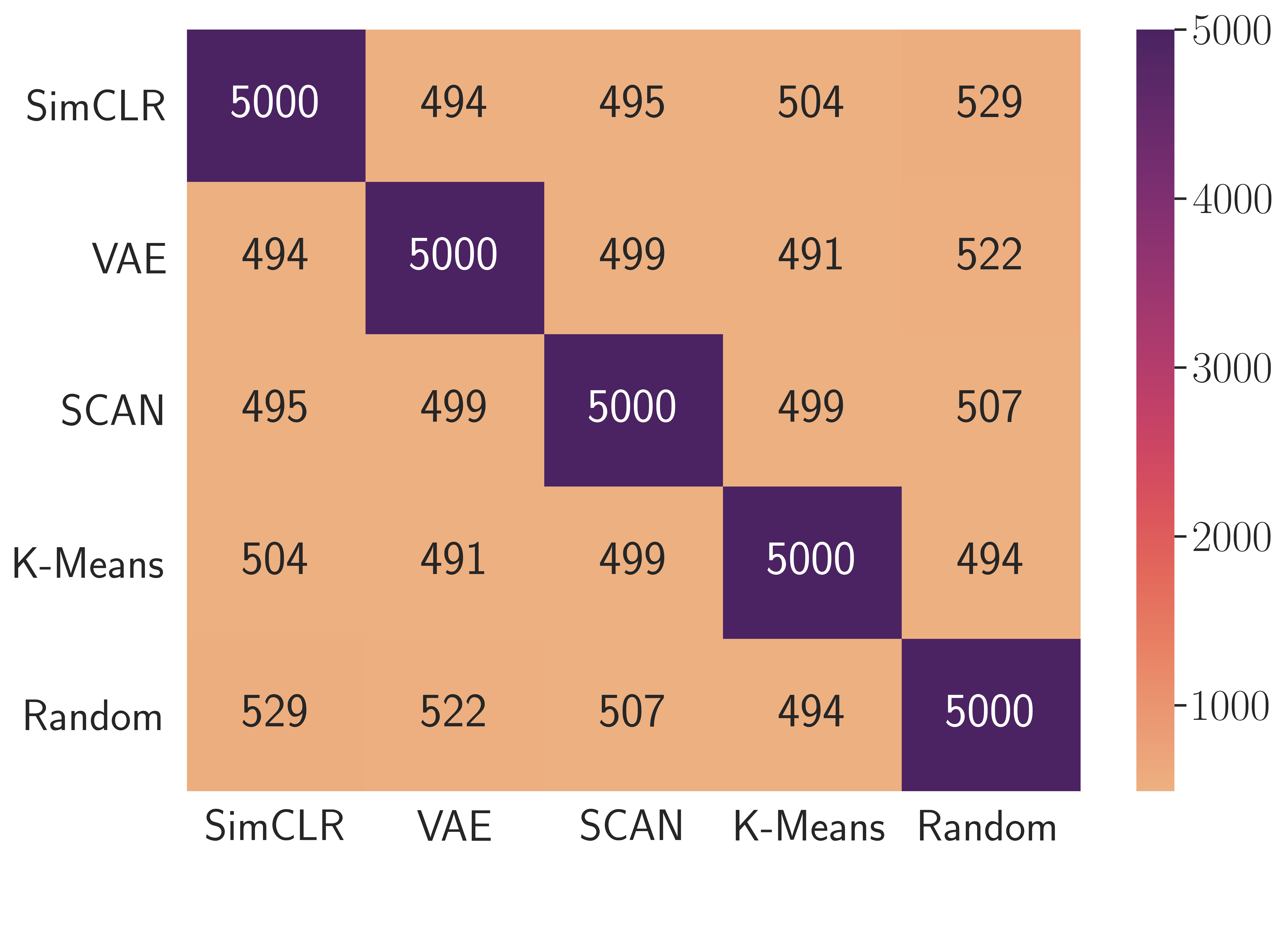} \\
\end{tabular}
\caption{CIFAR-10: Initial pools visualized using t-SNE.}
% \vspace{-5pt}
\label{fig:tsne}
\end{figure}

\textbf{Comparing the Initial Pools: }We used SimCLR representations to obtain t-SNE embeddings on all initial pools of a randomly chosen Max-Entropy (ME) experiment on CIFAR-10. The t-SNE plots of 5000 data points are shown in Figure \ref{fig:tsne}. Unsurprisingly, we see no apparent inconsistencies or anomalies in either class distribution or inter-class relationships across the four initial pools except for the VAE-sampled initial pool whose class distribution is noticeably different than the others. 

The other four initial pools are nearly identical. A confusion matrix with their overlap statistics, shown in Figure \ref{fig:tsne}, shows that all five initial pools roughly shared approximately 10\% of the data points among themselves. Even with nearly 90\% of unique data points, all four initial pools contributed to largely similar model generalization error (as seen in Max-Entropy graph of Figure \ref{fig:cifar10_graphs}). 

\begin{figure}[h]
\begin{tabular}{ccc}
\includegraphics[width=0.3\linewidth]{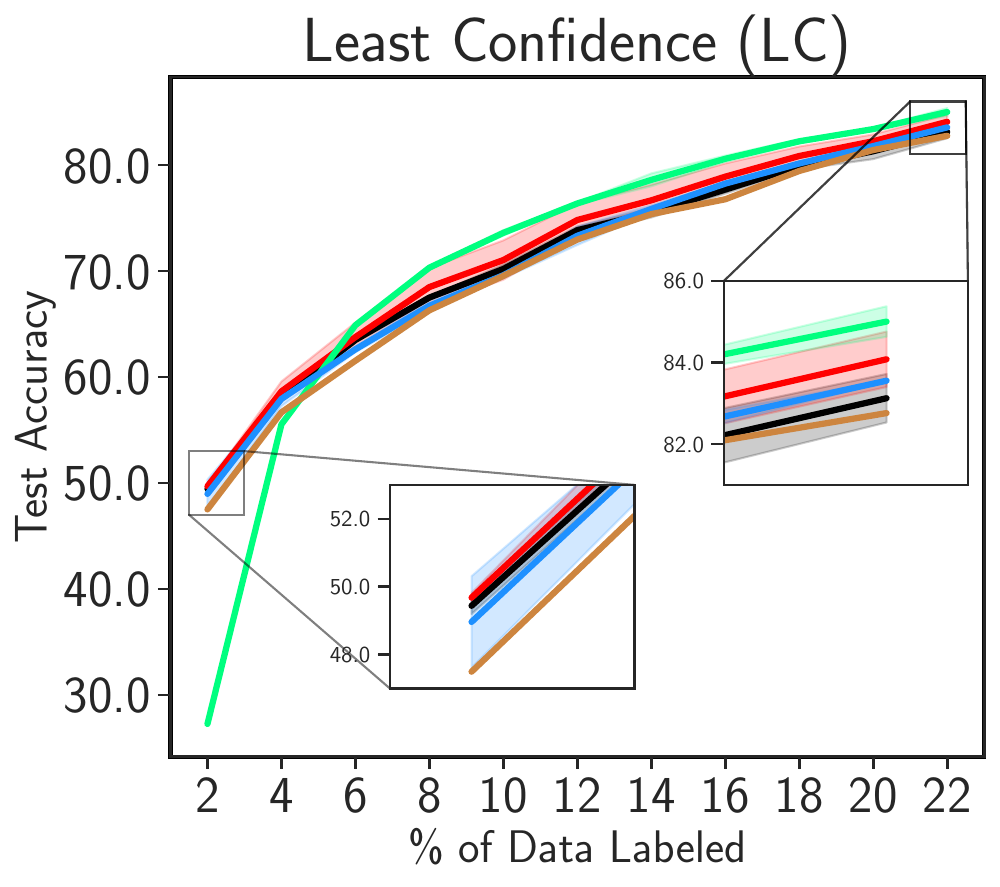} &
\includegraphics[width=0.3\linewidth]{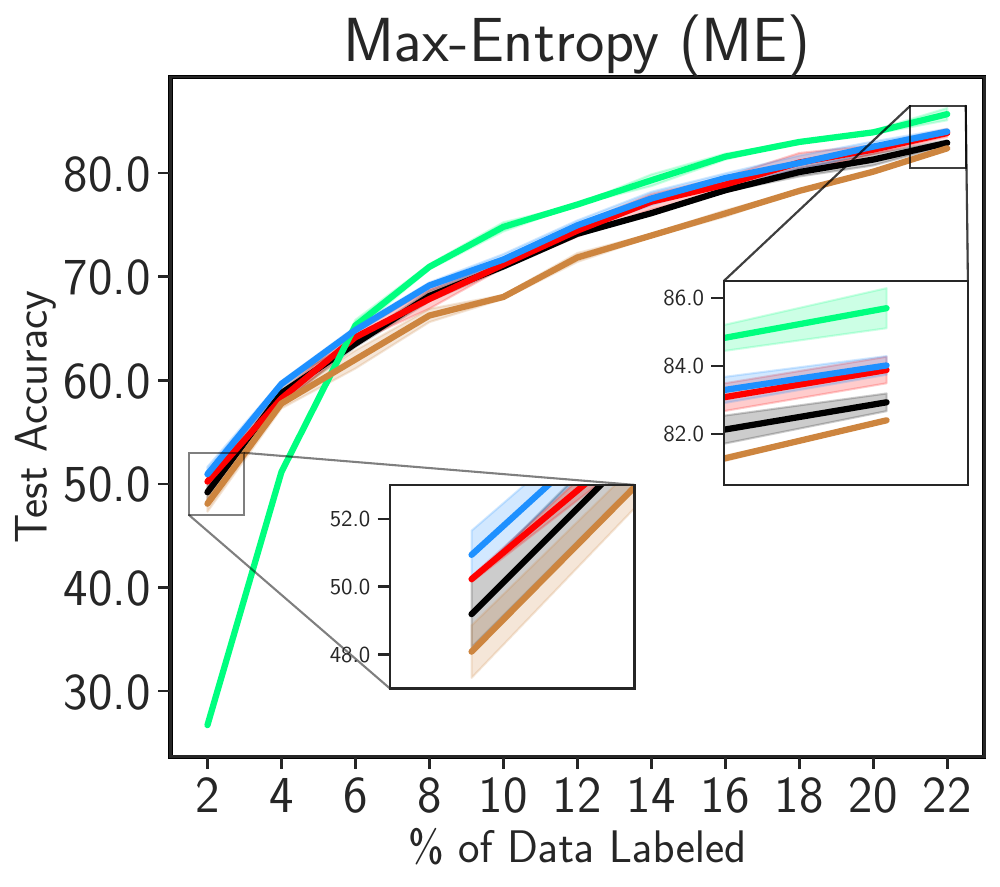} &
\includegraphics[width=0.3\linewidth]{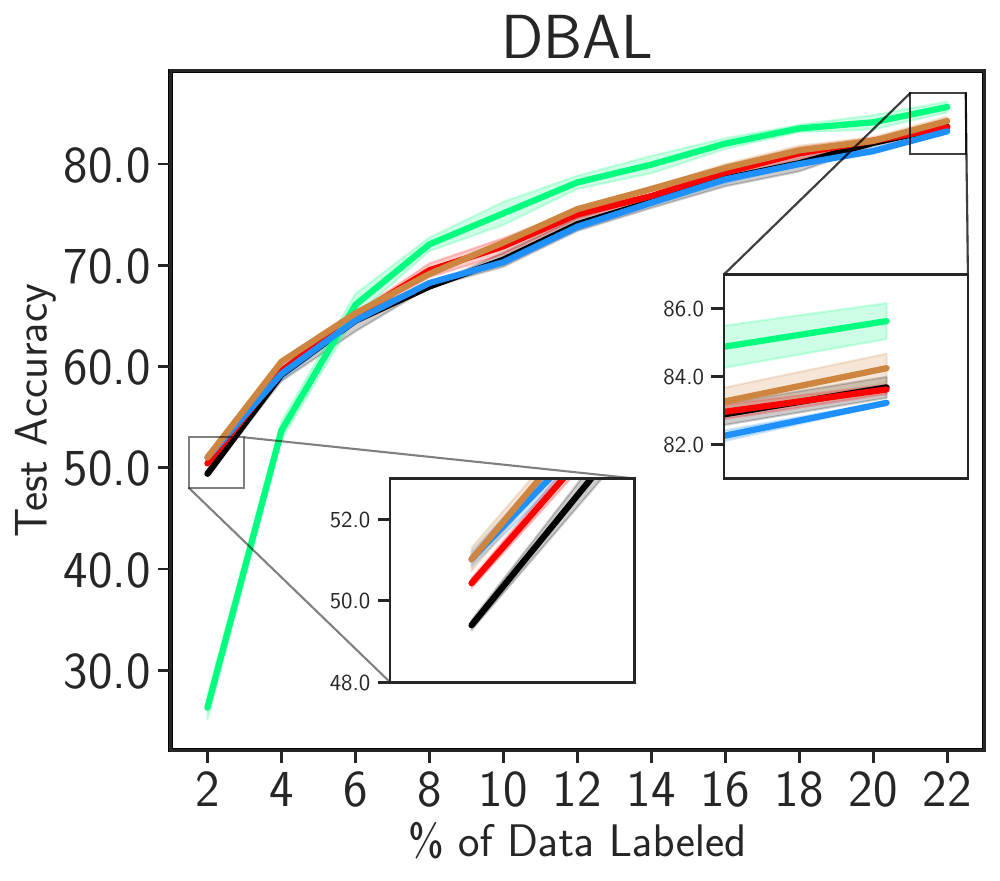}\\

\multicolumn{3}{c}{\includegraphics[width=0.7\linewidth]{figures/legend2.jpg}} \\
\end{tabular}
\caption{CIFAR-10: In low budget AL setting, only VAE initial pools show marginal performance gains over random initial pools.}
% \vspace{-5pt}
\label{fig:small_cifar10}
\end{figure}

\vspace{3mm}
\noindent \textbf{Low-Budget AL: } Is 10\% of data points too many for the model? Is that why we are unable to spot any potential performance differences between the four mostly unique initial pools? To find out if a low query budget can help spot performance differences, we repeated our experiments on CIFAR-10 for Max-Entropy (ME), Least Confidence (LC) and Deep Bayesian (DBAL) AL query methods but with just 1000 samples (2\% of the overall dataset size) in the initial pool. We set the AL budget to 1000 and allowed the AL cycles to run up to 10 episodes (22\% of the overall dataset size). The results of these experiments (averaged over 2 runs) are shown in Figure \ref{fig:small_cifar10}. All three AL methods benefit from VAE-sampled initial pools, albeit marginally, while other initial pools do not contribute to any performance gain compared to random initial pools.

\begin{figure}[h]
\begin{tabular}{ccc}
\includegraphics[width=0.3\linewidth]{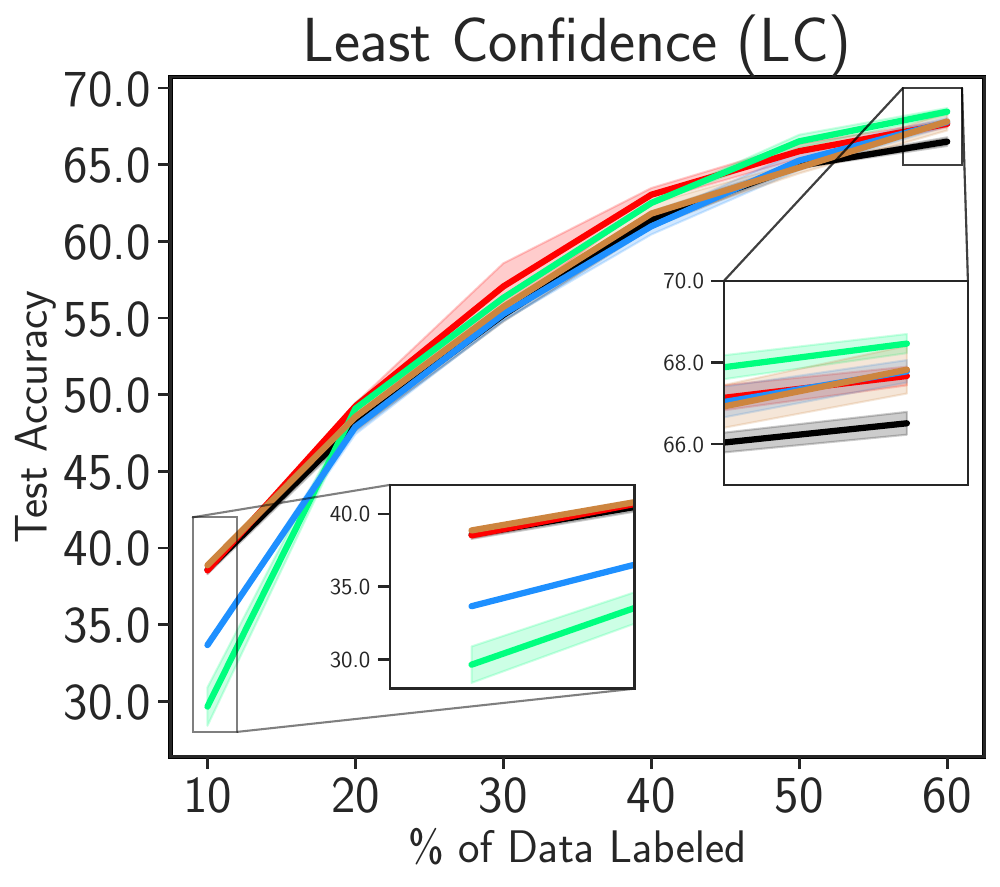} &
\includegraphics[width=0.3\linewidth]{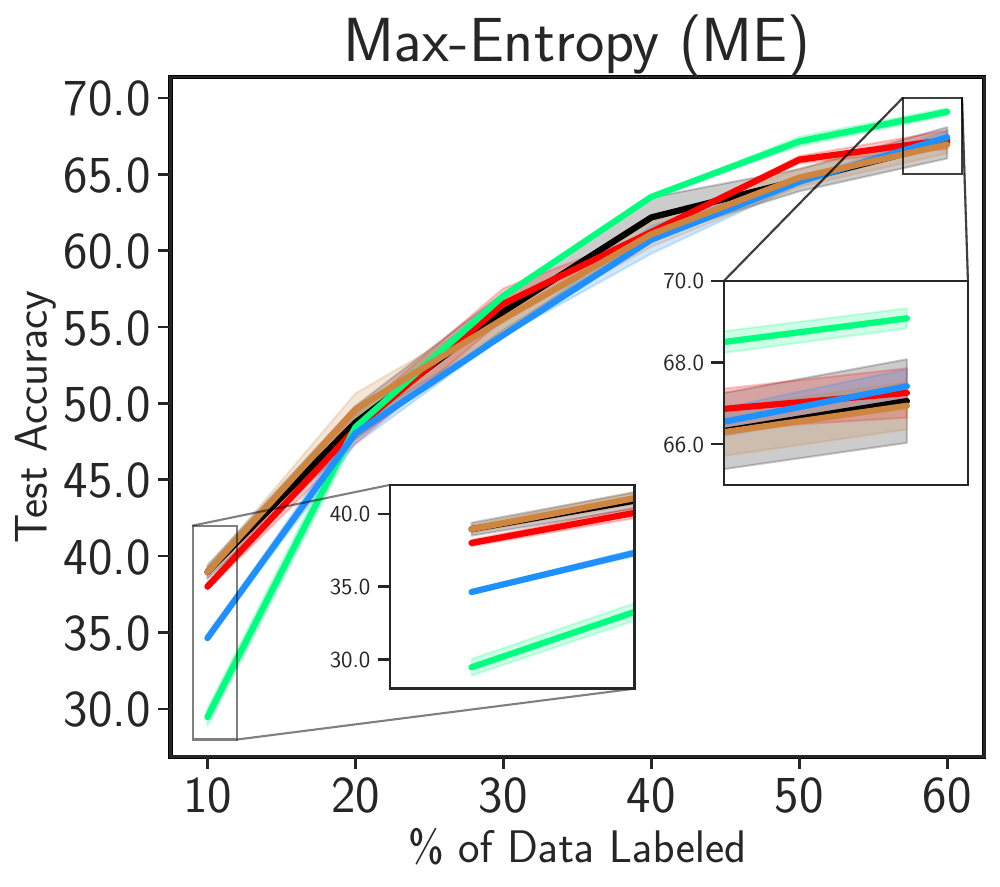} & 
\includegraphics[width=0.3\linewidth]{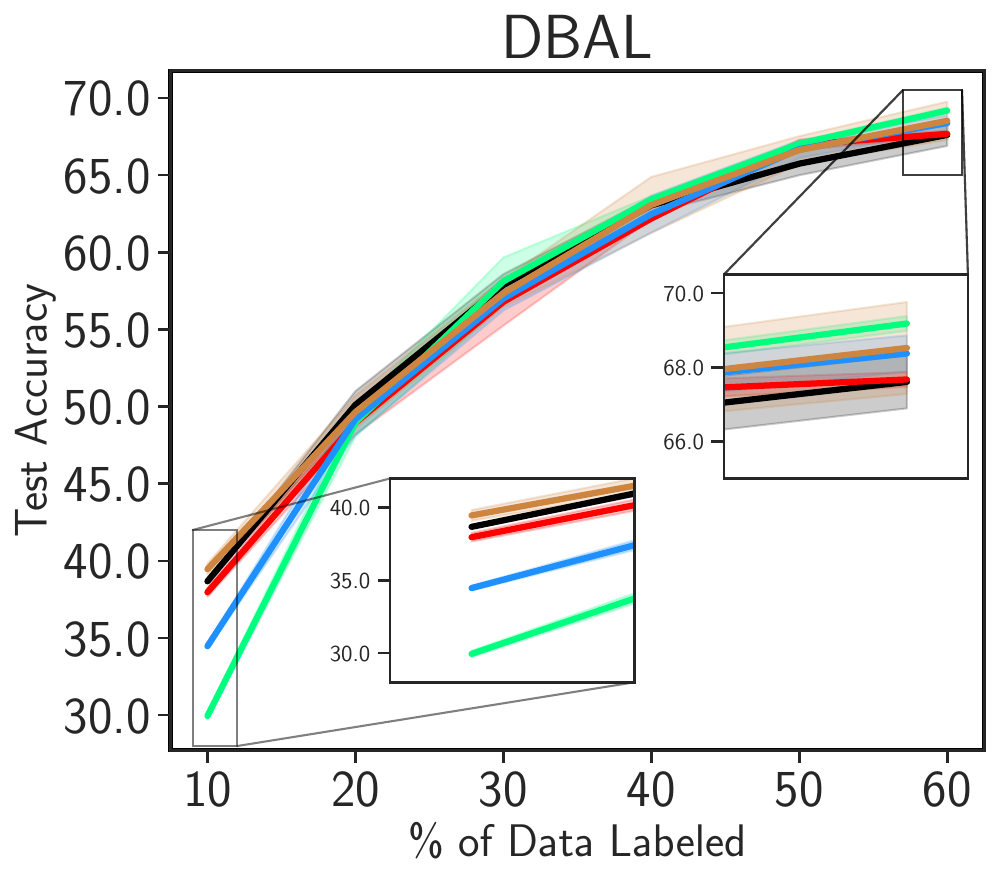} \\
\multicolumn{3}{c}{\includegraphics[width=0.7\linewidth]{figures/legend2.jpg}} \\
\end{tabular}
\caption{Long-Tail CIFAR-10: Our unsupervised sampling methods (SCAN and K-Means), motivated by this imbalance setting, did not improve LC, ME, DBAL query method performances. In the long run, VAE-based initial pools show marginal performance gains over random initial pools.}
% \vspace{-5pt}
\label{fig:imbalanced_cifar10}
\end{figure}

\vspace{3mm}
\noindent \textbf{Long-Tail CIFAR-10: }One of the motivations behind our proposed unsupervised method (Section \ref{sec:unsupervised}) was to allow AL cycles to start with a balanced initial pool, which spans the entire data distribution, when dealing with imbalanced datasets. To that end, we created a Long-Tail CIFAR-10 with an imbalance factor of 50 (\cite{class_imbalance_Cui2019}). We report the experiment results on three AL methods (ME, LC, DBAL) averaged over 2 runs in Figure \ref{fig:imbalanced_cifar10}. Surprisingly, our unsupervised initial pool sampling methods did not help the three AL methods. In fact, models trained on SCAN-based initial pools did consistently worse than models trained on random initial pools. Once again, VAE-based initial pools positively contribute to three AL methods albeit the performance gain is quite marginal.  

\section{More Training Details}\label{supp}
In this section we mention more training details necessary for reproducing our experiments.

\subsection{Slightly Modified ResNet18 Model}
To add an extra projection layer as the penultimate layer of the model is a convention in self-supervised learning methods (\cite{simlcr_Chen2020ASF, BootstrapYOGrill2020}). To be consistant with the ResNet18 model used for SimCLR and SCAN training, we added a projection layer to the model just before the final fully connected layer. Projection dimension was set to 128 for MNIST, CIFAR-10/100, 512 for Tiny ImageNet experiments. Also, the official ResNet18 implementation doesn't include any dropout layers in it. To allow for DBAL method to run Monte Carlo simulations, we added a dropout layer with \textit{p}=0.5 after the flattening layer. We only did this for DBAL experiments.   

\subsection{Hyperparmeters for AL Training}
For all experiments on all datasets, we set momentum = 0.9, \textit{wd} = 3$e^{-4}$, gamma = 0.1. Other hyperparameter choices are as follows:  

\begin{table}[h]
\centering
\begin{tabular}{|c|c|c|c|c|c|}
\hline
Dataset      & Epochs & Optimizer & Learning Rate & Scheduler              & Batch Size \\ \hline
CIFAR-10/100 & 200    & SGD       & 0.025         & Cosine (0.1) & 96         \\
MNIST        &  100      & Adam          & 0.005              &                        None &  64 \\
TinyImageNet & 100    & Adam      & 0.001         & None                   & 200        \\ \hline
\end{tabular}
\caption{Hyper-parameters of AL Cycles}
\label{table:hyperparams}
\end{table}

\subsection{SimCLR, SCAN and VAE Training}
For all four datasets we train SimCLR and SCAN with largely similar hyperparameters. We use the official implementation of SCAN\footnote{\url{https://github.com/wvangansbeke/Unsupervised-Classification/}} (includes SimCLR implementation as well) for training and use their recommended hyperparmeters across all experiments. In case of CIFAR-100, we follow the standard practice and group the 100 classes into 20 super classes before training SimCLR (the grouping details can be found in the official SCAN repository). Evaluation metrics of our final SimCLR + SCAN + Self Labeling models are as follows:

\begin{table}[h]
\centering
\begin{tabular}{|c|c|c|c|}
\hline
Dataset  & ACC & NMI & ARI  \\ \hline
CIFAR-10 & 0.70   & 0.46   & 0.38 \\
CIFAR-100 & 0.44   & 0.41   & 0.25 \\
MNIST        &  0.86  & 0.72  & 0.72 \\
Tiny ImageNet & 0.10  & 0.07  & 0.05 \\ \hline
\end{tabular}
\caption{Final Model Performances after Self-Labeling (SimCLR + SCAN + Self-Label)}
\label{table:simclr_final}
\end{table}

In case of VAE training, we trained a Vanilla VAE\footnote{\url{https://github.com/AntixK/PyTorch-VAE}} on the entire training data with hyperparameters as follows: optimizer = Adam, $lr = 0.001$, epochs = 100, momentum = 0.9, $wd = 5e^{-4}$, batch size = 200, for all four datasets. We used 5\% of the training data as the validation set. The model weights at epoch with best loss are saved for initial pool sampling.

\section{Conclusion}
In this paper, we proposed two kinds of strategies -- self-supervision based and clustering based -- for intelligently sampling initial pools before the use of active learning (AL) methods for deep neural network models. Our motivation was to study if there exist good initial pools that contribute to better model generalization and better deep AL performance in the long run. Our proposed methods and experiments conducted on four image classification datasets couldn't conclusively prove the existence of such good initial pools. However, a surprising outcome of this study was how initial pools sampled with a simple VAE task contributed to improved AL performance, better than more complex SimCLR and SCAN tasks. Even though VAE-based initial pools worked better than random initial pools only on one dataset (CIFAR-100), ablation studies on low budget CIFAR-10 settings as well as on Long-Tail CIFAR-10 point towards potential in VAE-sampled initial pools. Are images that are hard to reconstruct for VAEs good for generalization? Can better generative models like GANs do better than VAEs? We leave this for future work. While our methods and findings could not conclusively prove our hypothesis that AL methods can benefit from more intelligently chosen initial pools, we are optimistic about the potential this research direction holds.

\section*{Acknowledgements}
We thank DST, Govt of India, for partly supporting this work through the IMPRINT program (IMP/2019/000250). We also thank the members of Lab1055, IIT Hyderabad for engaging and fruitful discussions. Last but not the least, we thank all our anonymous reviewers for their insightful comments and suggestions, which helped improve the quality of this work.

\bibliography{chandra21}
\end{document}